\documentclass[lettersize,journal]{IEEEtran}
\usepackage{amsmath,amsfonts}
\usepackage{algorithmic}
\usepackage{algorithm}
\usepackage{array}
\usepackage[caption=false,font=normalsize,labelfont=sf,textfont=sf]{subfig}
\usepackage{textcomp}
\usepackage[table,xcdraw]{xcolor}
\usepackage{stfloats}
\usepackage{url}
\usepackage{verbatim}
\usepackage{graphicx}
\usepackage{booktabs}
\usepackage{bbding}
\usepackage{multirow}
\usepackage[accsupp]{axessibility}
\usepackage[table,xcdraw]{xcolor}
\usepackage{colortbl}
\usepackage{subfig}
\usepackage{pifont}
\usepackage[colorlinks=true,   
            linkcolor=red,     
            citecolor=blue,    
            urlcolor=magenta,  
            allbordercolors=white]{hyperref} 
\usepackage{orcidlink}
\begin{document}

\title{Compact Snapshot Spectral Imaging with Calibration-Free Aperture Diffraction}

\author{Tao Lv\orcidlink{0009-0006-8269-7623}, Quan Yuan, Shiqiao Li, Chenglong Huang\orcidlink{ 0009-0007-4552-7737}, Linsen Chen\orcidlink{0000-0002-1259-135X}, Chongde Zi, Shuming Wang, \\ Xun Cao\orcidlink{0000-0003-3094-4371},~\IEEEmembership{Member, IEEE}

\thanks{Tao Lv, Quan Yuan, Shiqiao Li, Chenglong Huang, Linsen Chen, Chongde Zi, Shuming Wang, and Xun Cao are with Nanjing University, Nanjing, 210023, China. E-mail: $\lbrace$lvtao, lishiqiao, chenglong-huang$\rbrace$ @smail.nju.edu.cn, $\lbrace$yq, chenls, zichongde, wangshuming, caoxun$\rbrace$ @nju.edu.cn.}
\thanks{Tao Lv and Quan Yuan contributed equally.}
\thanks{Shuming Wang and Xun Cao are the corresponding authors.}
}

\markboth{Journal of \LaTeX\ Class Files,~Vol.~14, No.~8, August~2021}%
{Shell \MakeLowercase{\textit{et al.}}: A Sample Article Using IEEEtran.cls for IEEE Journals}

\maketitle

\begin{abstract}
    Snapshot Spectral Imaging (SSI) provides high-dimensional temporal-spatial-spectral observation to uncover intrinsic physical characteristics. However, its complex system and repetitive calibration requirements hinder edge applications. Here, we propose a compact, cost-effective, calibration-free SSI method, Aperture Diffraction Imaging Spectrometer (ADIS), which consists only of a diffractive lens with a binary mask and a Bayer-filtered sensor, requiring no additional physical footprint compared to standard RGB cameras. ADIS disperses and multiplexes wavelengths, mapping energy to distinct sensor locations, enabling full-resolution recovery from superpixel-level encodings. 
    ADIS directly leverages theoretically computed PSFs to enable calibration-free spectral reconstruction, while tolerating lens-dependent variations across different optical configurations and bridging the gap between simulation and reality.
    To achieve SSI by solving a sparsely-constrained inverse problem, we introduce the Orthogonal Diffraction-Aware Unfolding Framework (ODAUF) with Voxel Shift Transformer (VST) for improved orthogonal diffraction perception. Integrating VST into ODAUF forms the efficient Orthogonal Diffraction-Aware Unfolding Voxel Shift Transformer (ODAUVST), delivering excellent recovery and reduced parameters. By elaborating on theory, systematic and comprehensive comparing, and demonstrating real SSI results, we validate the superiority of ADIS, achieving calibration-free full-resolution SSI within a commercial camera footprint.

\end{abstract}


\begin{figure*}[t]
    \begin{center}
    \includegraphics[width=1\linewidth]{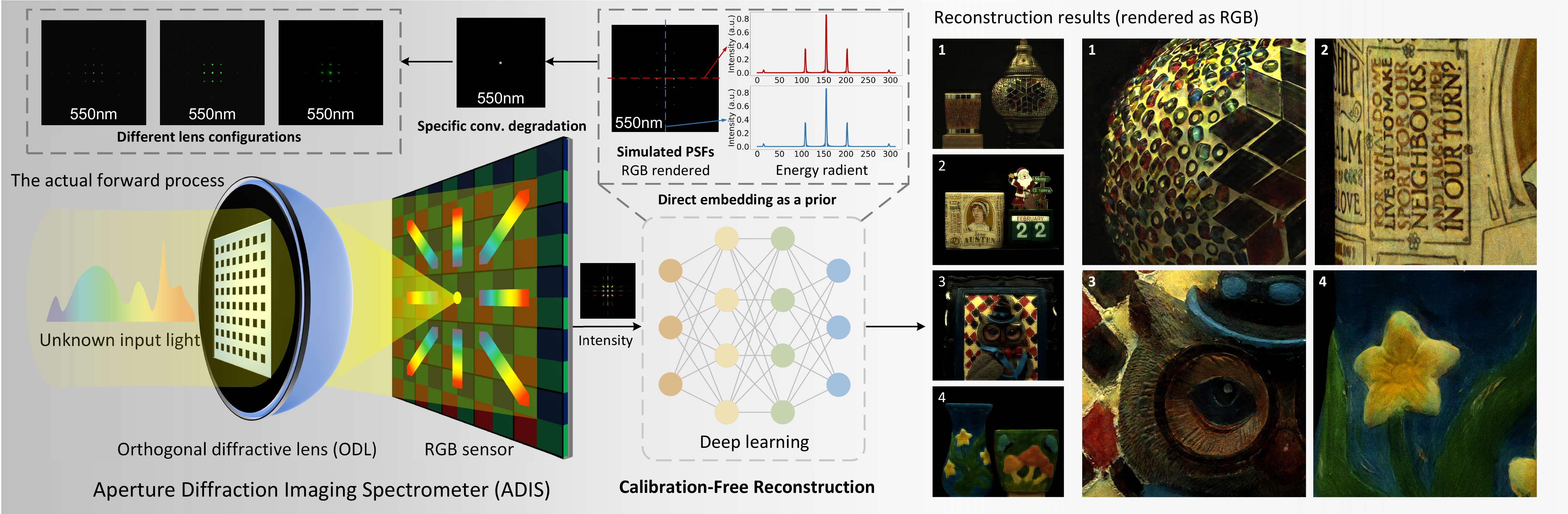}
    \end{center}
\vspace{-0.3cm} 
\caption{The imaging principle of ADIS. Light from the target scene is modulated by the Orthogonal Diffraction Lens (ODL) and captured by an array-filtered image sensor. The compressed measurements are reconstructed into HSIs using theoretically computed PSFs and filter functions (sensor parameters) as priors. The theoretically computed PSFs enable calibration-free reconstruction while tolerating lens-dependent variations across different optical configurations. The right side shows exemplar measurements and the corresponding reconstructed HSIs.}
\label{fig1}
\vspace{-0.5cm} 
\setlength{\belowcaptionskip}{-0.5cm}
\end{figure*}

\section{Introduction}

Hyperspectral imaging has found extensive applications across various fields due to its inherent optical characteristics, which enable the detailed identification of physical and chemical properties. By sacrificing temporal resolution, hyperspectral imaging captures fine-grained spectral data at each spatial location, producing a hyperspectral image (HSI) that provides a rich representation of the spectral features in the observed scene. This additional spectral information is crucial for distinguishing materials with the same color but different compositions, thus enabling the differentiation of their physical and chemical properties. Snapshot spectral imaging (SSI) advances this capability further, aiming to capture the high-dimensional spectral description of a scene in a single exposure, thereby extending the perceptual capabilities beyond the human eyes~\cite{cao2016computational}. It has found significant applications in fields such as industrial monitoring~\cite{hagen2020survey}, astronomical observation~\cite{cappellari2016structure}, combustion dynamics~\cite{he2020compressed}, and cellular dynamics~\cite{taruttis2015advances}.

However, traditional SSI systems rely on specific optical components, such as dispersive optics (prism or grating), occlusion masks, multiple relay lenses, and an imaging lens, resulting in a large and complex form factor, limiting their practical application at the edge~\cite{PMVIS,SDCASSI, DDCASSI, CTIS}. Furthermore, the accurate characterization of dispersion and spatial encoding in such systems requires precise calibration, which must be repeatedly performed whenever the systems are disturbed~\cite{lv2024efficient}.

In summary, the application of SSI at the edge faces two key limitations: (1) large and complex form factor; (2) the necessity for frequent recalibration. Although recent efforts have been made to miniaturize SSI systems based on diffractive optical elements (DOEs), calibration is still required to account for the modulation effects of the DOE across different spectral bands due to manufacturing challenges~\cite{DOE_Jeon,wang2024non}. Despite meticulous calibration, the simulation-to-reality (sim2real) gaps often persist, inevitably introducing artifacts in reconstructed outputs.

An alternative approach, hyperspectral super-resolution from RGB, presents another solution, but ensuring fidelity in spectral reconstruction of unknown scenes remains challenging, even though performance at the dataset level is acceptable. Other methods, such as utilizing metalenses~\cite{yesilkoy2019ultrasensitive}, photonic crystals~\cite{wang2019single}, and Fabry-Pérot based random filters~\cite{yako2023video}, leverage micro- and nanofabrication techniques to optimize filtering and encoding~\cite{miniaturization}. However, from the perspectives of manufacturing costs and sample consistency, these approaches remain largely confined to laboratory settings and are not yet scalable for broader applications.

To address the challenges of existing SSI methods, we propose a cost-effective, compact, and user-friendly SSI solution, \textbf{A}perture \textbf{D}iffraction \textbf{I}maging \textbf{S}pectrometer (ADIS). In ADIS, the lens features a tightly fitting orthogonal mask, uniformly distributed in binary form along the orthogonal direction. This configuration forms an \textbf{O}rthogonal \textbf{D}iffraction \textbf{L}ens (ODL), which can be rigorously characterized physically. The ODL enables SSI without increasing the camera's physical footprint and eliminates the need for calibration. Furthermore, ADIS directly utilizes theoretically computed point spread functions (PSFs) for forward modeling and reconstruction guidance, while accommodating lens-dependent convolutional degradation arising from different optical configurations. By explicitly tolerating such practical optical variations, ADIS further strengthens its calibration-free property.

In the aperture plane, the orthogonal mask modulates the incident light amplitude and generates lattice-like diffraction PSFs (Fig.~\ref{fig1}), which multiplex spectral information of different wavelengths onto distinct sensor locations and enable full-resolution spectral recovery from superpixel-level filtered measurements.
Additionally, according to Babinet's principle, complementary masks produce similar diffraction patterns, which can be exploited to enhance the energy efficiency. However, the orthogonal diffraction encoding presents a challenging inverse reconstruction problem.
To address this, we propose an Orthogonal Diffraction-Aware Unfolding Framework (ODAUF), which possesses decoupled perceptual capabilities for diffraction degradation. Within this framework, we have meticulously designed a diffraction-aware enhanced Voxel Shift Transformer (VST) that implements high-dimensional voxel orthogonal shifts through shuffle and fractional pixel shift operations. The VST serves as the executor for the ODAUF's regularization component.

Notably, we investigate the specific implementation of the VST core component, the Learning Shifts Module (LSM), to enhance generalizability and improve reconstruction quality in both high-resolution simulations and real-world experiments.
By integrating the VST into the ODAUF, we establish a Transformer-based deep unfolding algorithm, the Orthogonal Diffraction-Aware Unfolding Voxel Shift Transformer (ODAUVST), for ADIS reconstruction. This integration enables ADIS to achieve optimal high-dimensional spectral acquisition performance within a compact SSI configuration.

To validate the imaging performance of ADIS, we systematically analyze the imaging optical theory and reconstruction algorithms, comparing various imaging systems and hardware design parameters, demonstrating real spectral imaging results with single-exposure captures. Overall, we have achieved full-resolution SSI without system calibration, effectively extending the spectral perception capabilities of digital cameras within the footprint of standard RGB cameras.
Specifically, our contributions are as follows:

$\bullet$ Development of a cost-effective SSI system that integrates an orthogonal diffraction lens (ODL) with a bare sensor, enabling calibration-free and sim2real-aligned compact SSI.

$\bullet$ Introduction of a principled diffraction-projection-guided Framework, ODAUF, based on the unfolding architecture of an optimization procedure with a spatial-spectral prior, mitigating the ill-posedness of spectral reconstruction in ADIS.

$\bullet$ Design of the diffraction-aware enhanced VST, leveraging fractional shift operations to enhance degradation perception and generalizability. By plugging VST into ODAUF, we realize ODAUVST with excellent restoration and reduced parameters in high-resolution simulations and real experiments. 

$\bullet$ A prototype device combined with ODAUVST is constructed for calibration-free SSI, while extensive experiments demonstrate the excellent superiority of our proposed methods.


\begin{figure*}[t]
    \begin{center}
    \includegraphics[width=1\linewidth]{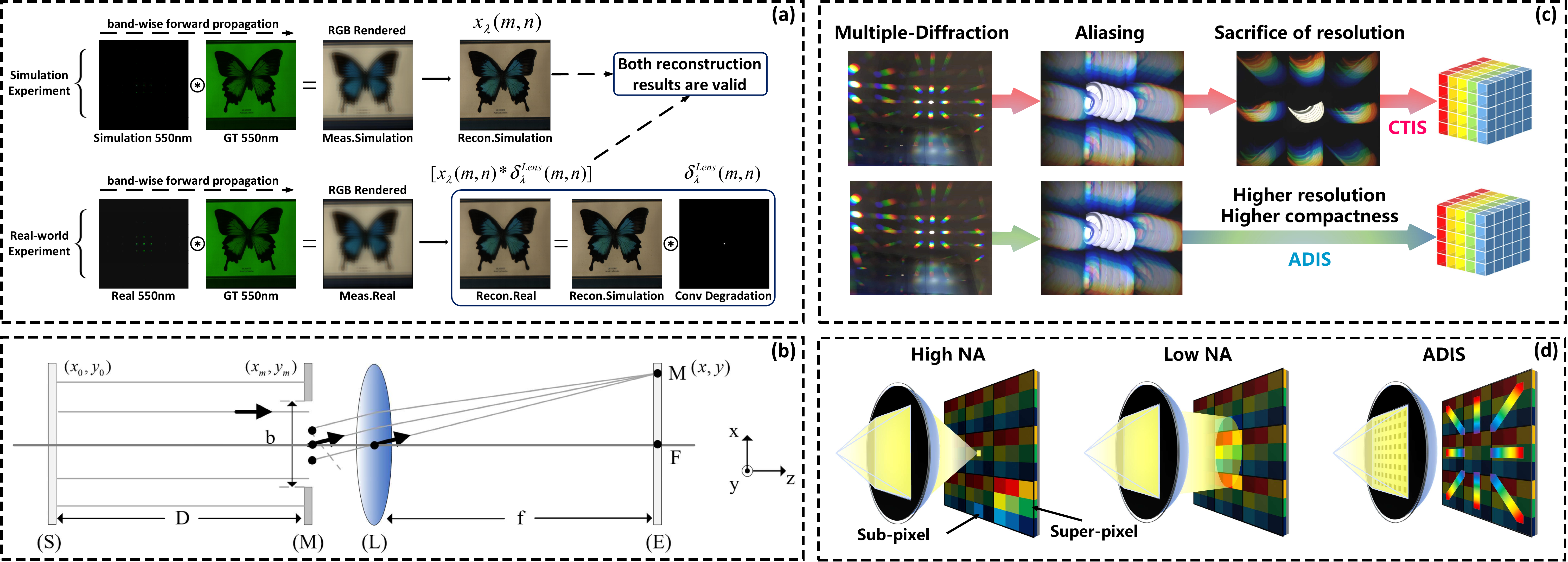}
    \end{center}
\vspace{-0.5cm} 
\caption{(a) Calibration-free reconstruction of ADIS using theoretically computed PSFs, with robustness to lens-dependent convolutional degradation across different lens configurations; (b) ADIS forward propagation with collimated illumination; (c) illustrates the principles of CTIS and strategy of using long optical path by sacrificing spatial resolution, while ADIS reconstructs from aliasing; (d) depicts compact imaging methods with mosaic filter sensors.}
\label{fig3}
\vspace{-0.3cm} 
\setlength{\belowcaptionskip}{-0.5cm}
\end{figure*}

\section{Related Work}

SSI aims to achieve high-dimensional spectral data acquisition in a single exposure. 
In the past decades, SSI was exemplified by integral field spectroscopy (IFS)~\cite{IFS}. While IFS achieves a trade-off between spectral and spatial resolution, its high hardware cost has hindered adoption beyond laboratory settings.
With the advancement of computational imaging, computational-based SSI techniques have evolved, incorporating approaches such as spatial encoding, array-filtered encoding, and aperture encoding. Here, we provide a comprehensive analysis of these computational SSI designs and corresponding reconstruction algorithms.

\vspace{-0.2cm}
\subsection{Spatial Encoding Methods}

A typical spatial encoding method utilizes an occlusion mask positioned in the image plane, exploiting the sparsity of spatial information from the scene to encode the spatial domain. This is followed by the use of dispersive elements to create explicit spectral encoding. Representative techniques such as CASSI~\cite{SDCASSI,DDCASSI} and PMVIS~\cite{PMVIS, HVIS}, have been developed based on this approach. However, due to the complexity of the system and calibration errors, these techniques are largely confined to laboratory settings. A common characteristic of spatial encoding methods is the necessity of a $4f$ structure to provide an additional image plane for spatial gating, while dispersive optics operate on the object plane. 

As a result, the spatial encoding paradigm is inherently difficult to implement in compact systems.
Although some modified approaches, such as CCASI~\cite{correa2014compressive, arguello2014colored, rueda2015multi,correa2015snapshot} and DCSI~\cite{lin2014dual}, have attempted to design systems with more randomized sensing matrices or enhanced optical throughput~\cite{cao2016computational, cao2021hyperspectral}. They have not significantly reduced system complexity or calibration requirements.
In contrast, our approach integrates convolutional and filter-based encoding within a compact $2f$ system architecture, enabling a calibration-free SSI system within the footprint of a standard RGB camera.

\vspace{-0.2cm}
\subsection{Array-Filtered Encoding Methods}

Array-filtered encoding methods typically recover the desired channels by utilizing tiled spectral filter arrays, which feature a unique arrangement of superpixels periodically arranged on the sensor plane. However, as the number of sampled channels increases, spatial resolution diminishes~\cite{lapray2014multispectral}. Although some demosaicking techniques can be integrated with filter-array-based methods, they do not primarily rely on data that is originally captured by the sensor~\cite{mihoubi2017multispectral}.
 
Additionally, due to material and design limitations, array-filtered encoding methods often cannot be easily adapted to meet specific application needs with simple adjustments.
Of course, despite the limitations of detector and filter sizes, spectrometers based on specific filters offer clear advantages in terms of miniaturisation~\cite{miniaturization}. Various design solutions, such as thin films~\cite{wang2007concept}, photolithographic coatings~\cite{bian2024broadband}, Fabry-Pérot random filters~\cite{yako2023video}, planar photonic crystals~\cite{wang2019single}, and metasurfaces~\cite{yesilkoy2019ultrasensitive}, have been demonstrated in laboratory settings for the development of array-filter-based spectrometers. 
However, from the perspectives of manufacturing cost and sample consistency, these approaches are not yet scalable for broader applications. In contrast, our method combines commercial array filters, such as the Bayer array, with convolutional encoding of an ODL, enabling cost-effective SSI.

\vspace{-0.2cm}
\subsection{Aperture Encoding Methods}

Aperture encoding methods typically use dispersive optics at the aperture plane to spectrally encode the scene. A representative example of this paradigm is the Computational Tomography Imaging Spectrometer (CTIS)~\cite{CTIS}, which employs a combination of gratings with different orientations to achieve combined measurements of linear projections within a constrained spatial range. The occlusion mask and complex spectral projections result in a system with a complex structure and the need for repetitive calibration.
The complexity of the CTIS arises from minimizing data degradation by sacrificing spatial resolution, but advancements in computational imaging now allow recovery of high-fidelity, high-dimensional data from highly degraded measurements, thereby enabling a reduction in system complexity.
Some methods use a single dispersion to blur the scene, but leading to a highly ill-conditioned problem and low reconstruction accuracy because the spectral encoding is only at the edges of the objects in the scene~\cite{Aprism}. To further increase system compactness, DOEs are employed to reconstruct 3D HSIs based on sparsity assumptions~\cite{compact, shi2024learned, shi2024split}. Nonetheless, the robustness of modulation in these methods remains affected by manufacturing errors and environmental disturbances during use, necessitating high-precision and repetitive calibration. 
In contrast, ADIS leverages a uniformly distributed binary mask to create diffraction patterns similar to those in CTIS, while uniformly capturing diffraction aliasing (Fig. \ref{fig3}). This minimalist design allows the system to operate within the footprint of a standard commercial camera, eliminating the need for calibration and significantly enhancing the potential for SSI applications at the edge.

\begin{figure*}[t]
    \begin{center}
    \includegraphics[width=1\linewidth]{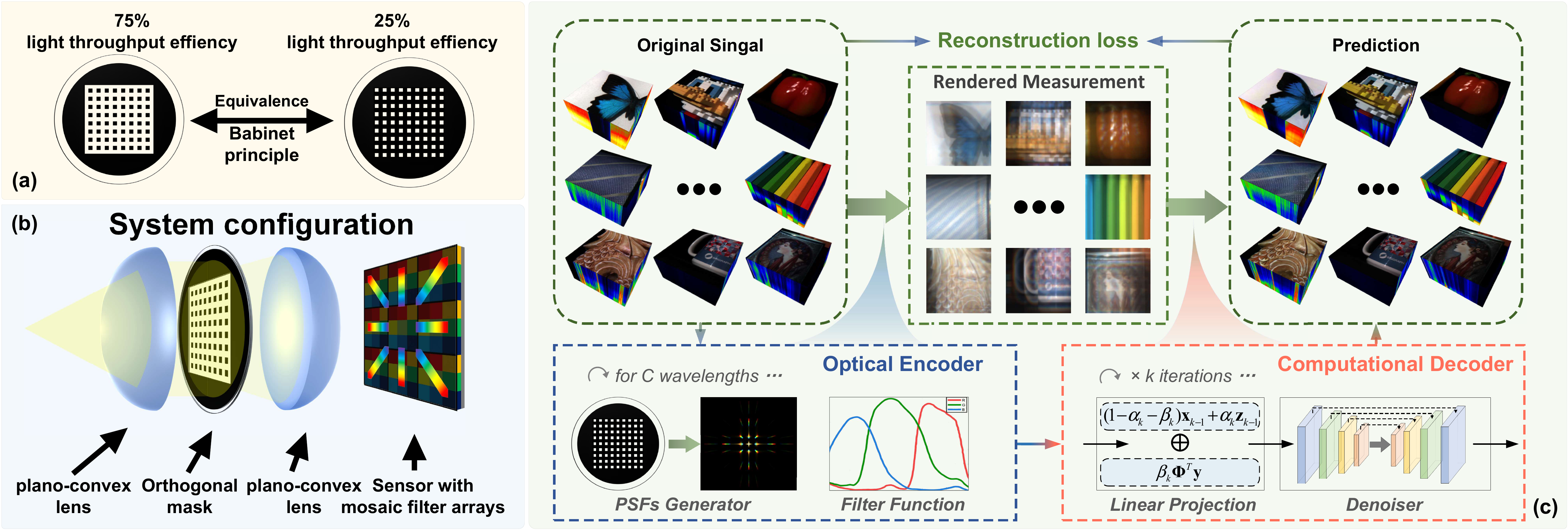}
    \end{center}
\vspace{-0.3cm} 
\caption{(a) demonstrates the application of the Babinet principle in the design of ADIS masks, which can be used to enhance light throughput; (b) depicts an alternative configuration of ADIS whereby the mask is located between two plano-convex lenses, producing a similar encoding process; (c) is the overview of proposed Aperture Diffraction Imaging Spectrometer (ADIS). The system consists of an optical encoder and a computational decoder, which encodes the HSIs into 2D images and reconstructs the underlying HSIs from the compressed measurements, respectively.}
\label{fig2}
\vspace{-0.3cm} 
\end{figure*}

\vspace{-0.2cm}
\subsection{Spectral Reconstruction}
Traditional iterative reconstruction approaches encounter significant challenges in terms of the time-consuming encoding process and the requirement for prior knowledge~\cite{yuan2016generalized,boyd2011distributed}. To address these challenges, end-to-end deep-learning approaches have been proposed and have demonstrated remarkable potential in optimizing complex ill-posed problems in various SSI systems~\cite{snapshot,TSA-net,l-net,HDnet}. Notably, $\lambda$-net~\cite{l-net} and TSA-net~\cite{TSA-net} have proposed dual-stage generative models and self-attention, respectively, to map HSIs from a single-shot measurement while modeling spatial and spectral correlation with reasonable computation cost. Recently, Transformer-based methods~\cite{MST,CST} have emerged as superior alternatives, outperforming previous methods and greatly improving reconstruction efficiency, but fail to fully leverage the perceptual structure of the imaging process. 
Accordingly, the physically guided deep unfolding methods, which effectively combine the iterative method with deep learning, achieve the desired results on specific image restoration problems~\cite{DAUHST,dong2018denoising,DNU}. To fully incorporate the physical prior of ADIS into the imaging inverse process, we propose a deep unfolding framework based on ADIS, optimizing the regularizer with orthogonal diffraction features for improved ADIS reconstruction performance.

\section{Aperture Diffraction Model}

This section introduces the proposed SSI system, ADIS, covering its basic configuration and diffraction principles. We also discuss design trade-offs of system parameters and analyze the system's robustness to external perturbations.

\vspace{-0.2cm}
\subsection{System Configuration}

Fig.~\ref{fig2} illustrates the configuration of our aperture diffraction imaging spectrometer (ADIS), which consists of only a diffractive lens and a commercial RGB camera. The diffractive lens consists of a normal lens and an orthogonally uniformly distributed binary mask that fits closely in front of the lens. In the experiment, the diffractive lens can be constructed using two plano-convex lenses and orthogonal masks, as shown in Fig.~\ref{fig2}(b), while sensors can incorporate more complex filtering configurations. 
When an object point with a smooth reflectance distribution is captured, ADIS disperses its spectral information across spatial locations in an orthogonal pattern. This pattern guides the spectral components to distinct sub-pixel positions on the mosaic filter encoding array. Consequently, each sub-pixel on the sensor captures different spectral bands from different spatial positions, enabling SSI with sub-superpixel resolution.

\vspace{-0.2cm}
\subsection{Diffraction Model}
\label{sec3.2}

The maintenance of spatial invariance in optical systems is an indispensable characteristic for effectively addressing interference-related issues. 
Here we discuss the depth invariance of ADIS. 
Suppose a monochromatic incident wave field $u_0$ with amplitude $A_0$, phase $\phi _0$ passes through the ODL: 

\begin{equation}
\setlength{\abovedisplayskip}{0.1cm}
\setlength{\belowdisplayskip}{0.1cm}
\label{1-7}
    {u_0}({x_m},{y_m}) = {A_0}({x_m},{y_m}){e^{i{\phi _0}({x_m},{y_m})}}
\end{equation}

An amplitude encoding and phase shift occurs by the ODL:
\begin{equation}
\setlength{\abovedisplayskip}{0.1cm}
\setlength{\belowdisplayskip}{0.1cm}
\label{1-8}
    {u_1}({x_m},{y_m}) = {u_0}({x_m},{y_m}){A_1}({x_m},{y_m}){e^{i{\phi _1}({x_m},{y_m})}}
\end{equation}

When the ADIS is illuminated by a point light source located at depth $Z$, the spherical wave field $u_0$ emitted by the source incident to the ODL can be represented by:
\begin{equation}
\setlength{\abovedisplayskip}{0.1cm}
\setlength{\belowdisplayskip}{0.1cm}
\label{11-9}
    {u_0}({x_m},{y_m};Z) \propto \frac{1}{\xi }{e^{ik(\xi  - Z)}}
\end{equation}

\noindent
where $\xi  = \sqrt {{x_m}^2 + {y_m}^2 + {Z^2}} $. Since the aperture size is negligibly smaller than the imaging depth, the following relationship exists: $\xi  \approx Z$. Then the wave field $u_1$ modulated by the ODL can be expressed as:
\begin{equation}
\setlength{\abovedisplayskip}{0.1cm}
\setlength{\belowdisplayskip}{0.1cm}
\label{11-10}
    {u_1}({x_m},{y_m};Z) \propto \frac{1}{Z}{A_1}({x_m},{y_m}){e^{i\{ k(\xi  - Z) + {\phi _1}({x_m},{y_m})\} }}
\end{equation}

Since $\xi  \approx Z$, the point source is relatively close to optical infinity, and $\left( {\xi  - Z} \right)  \ll {\phi _1}({x_m},{y_m})$ holds in Equation \ref{11-10}. Then Equation \ref{11-10} can be approximated as Equation \ref{1-8}. The above derivation verifies the depth invariance of the ADIS in a specific depth range.
This derivation confirms ADIS's depth invariance within a specific range. 

According to the Huygens-Fresnel principle, each point on a wavefront acts as a secondary source of spherical wavelets. Thus, we can treat each rectangular square aperture as a point source for a multi-slit diaphragm. Through the amalgamation of waves generated by each of point sources, we can effectively derive the wave pattern of the entire diaphragm:

\begin{equation}
\setlength{\abovedisplayskip}{0.1cm}
\setlength{\belowdisplayskip}{0.1cm}
    \label{1-1}
    {E_p} = {E_0}\frac{{\sin {\beta _1}}}{{{\beta _1}}}\frac{{\sin N{\gamma _1}}}{{\sin {\gamma _1}}}\frac{{\sin {\beta _2}}}{{{\beta _2}}}\frac{{\sin N{\gamma _2}}}{{\sin {\gamma _2}}}
\end{equation}

\noindent
where $\theta _1$ and $\theta_2$ are the diffraction angles in x- and y-directions respectively, ${\beta _1} = \frac{1}{2}kb\sin {\theta _1}$, ${\beta _2} = \frac{1}{2}ka\sin {\theta _2}$, ${\gamma _1} = \frac{1}{2}kd\sin {\theta _1}$, ${\gamma _2} = \frac{1}{2}kd\sin {\theta _2}$. Further, by utilizing the paraxial approximation in far-field imaging, the angular relationship can be transformed into a position relationship ($\sin {\theta _1} \approx \tan {\theta _1} = \frac{{{x}}}{{{f}}}$, $\sin {\theta _2} \approx \tan {\theta _2} = \frac{{{y}}}{{{f}}}$). $f$ denotes the distance between the diffraction array and the sensor. As a result, the intensity and position relationship of the diffraction pattern can be represented as follows:

\begin{gather}
\setlength{\abovedisplayskip}{-0.1cm}
\setlength{\belowdisplayskip}{-0.1cm}
\label{1-2}
   I(x,y,\lambda ) = {I_0} \cdot D(x,y,\lambda ) \cdot P(x,y,\lambda ) \\
   D(x,y,\lambda ) = \sin {c^2}(\frac{{b}}{{\lambda {f}}}{x})\sin {c^2}(\frac{{a}}{{\lambda {f}}}{y})\\
   P(x,y,\lambda ) = {\left[ {\frac{{\sin (N\frac{{\pi {d}}}{{\lambda {f}}}{x})}}{{\sin (\frac{{\pi {d}}}{{\lambda {f}}}{x})}}} \right]^2}{\left[ {\frac{{\sin (N\frac{{\pi {d}}}{{\lambda {f}}}{y})}}{{\sin (\frac{{\pi {d}}}{{\lambda {f}}}{y})}}} \right]^2}
\end{gather}

\noindent
where the formula $D(x,y,\lambda )$ is the diffraction factor describing the diffraction effect of each rectangular square hole. $P(x,y,\lambda )$ is the interference factor describes the effect of multi-slit interference and $({x},{y})$ denotes the spatial coordinates on the receiving screen.

Moreover, ADIS demonstrates resilience to $(x, y)$-direction device perturbations, providing that the modulation plane remains within the imaging optical path.
 Here, we assume a positional shift $p$ in the y-direction for the mask. Then we can get:
$ {E_p} = {E_0}\frac{{\sin {\beta _1}}}{{{\beta _1}}}\frac{{\sin N{\gamma _1}}}{{{\gamma _1}}}\frac{{\sin {\beta _2}}}{{{\beta _2}}}\frac{{\sin N{\gamma _2}}}{{{\gamma _2}}} \cdot {e^{ikp\sin \theta }}$. Taking the amplitude of the electric field, we can obtain $\left| {{E_p}} \right| = {E_0}\frac{{\sin {\beta _1}}}{{{\beta _1}}}\frac{{\sin N{\gamma _1}}}{{{\gamma _1}}}\frac{{\sin {\beta _2}}}{{{\beta _2}}}\frac{{\sin N{\gamma _2}}}{{{\gamma _2}}}$, which verifies the modulation robustness of the system.

\subsection{Hardware Parameter Selection and Logic}
\label{sec3.3}

\begin{figure}[t]
    \begin{center}
    \includegraphics[width=1\linewidth]{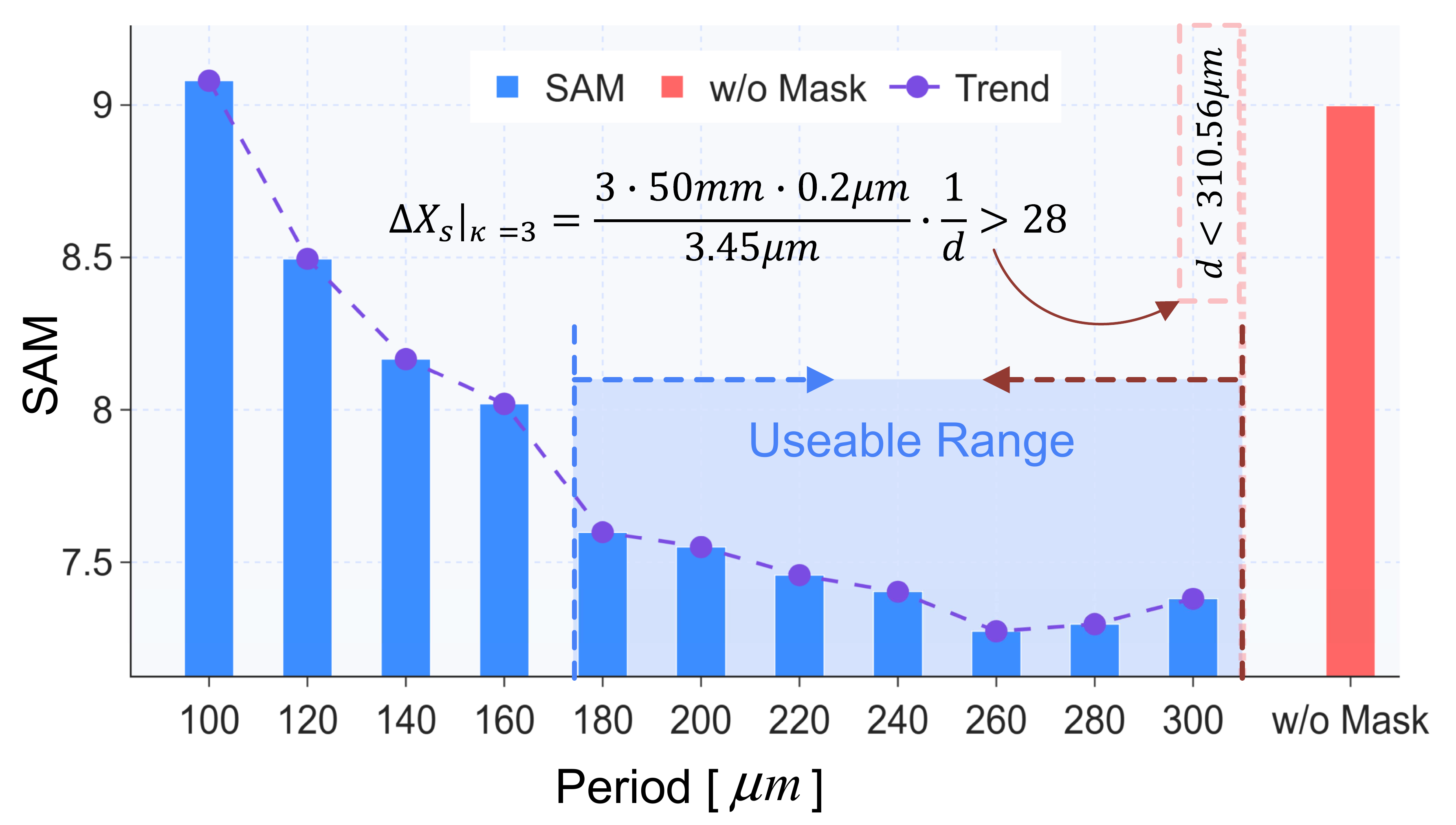}
    \end{center}
\vspace{-0.3cm} 
\caption{Comparison of reconstruction fidelity for masks with different periods on the test set, evaluated using Spectral Angle Mapping (SAM) as the spectral accuracy metric. Fifteen diverse outdoor scenes from the NTIRE 2020 dataset are used as a test set to assess the suitability of the chosen mask parameters.}
\label{Fig4}
\vspace{-0.3cm} 
\end{figure}

(1) Constraint from Target Spectral Resolution.

For a given spectral bandwidth $R = ({\lambda _{\max }} - {\lambda _{\min }})$ and target spectral resolution $\Delta \lambda$, the number of detector pixels over which a diffracted order is dispersed should be sufficiently large to resolve the desired spectral features. Denoting the sensor pixel pitch by $s_{pixel}$, the dispersion in pixel units $\Delta X_s$ for a given diffraction order  ${\kappa}$ can be approximated as:

\begin{equation}
\setlength{\abovedisplayskip}{0cm}
\setlength{\belowdisplayskip}{0cm}
\label{1.5.1}
    \Delta {X_s} = \frac{{\kappa  \cdot f \cdot ({\lambda _{\max }} - {\lambda _{\min }})}}{{d \cdot {s_{pixel}}}}
\end{equation}

To achieve the target resolution under a practically detectable signal level, we require $\Delta {X_s} > \frac{R}{{\Delta \lambda }}$. In this study, we focus on the third diffraction order, where even orders are suppressed and higher orders exhibit negligible energy. Given that our reconstruction covers 28 spectral bands across the $450 - 650 nm$ range, the dispersed stripe of the resolvable diffraction order must span at least 28 pixels to adequately sample the spectrum: ${\left. {\Delta {X_s}} \right|_{\kappa  = 3}}  > 28 \Rightarrow d < 310.56\mu m$.

(2) Trade-off between Inverse Problem Solving and Reconstruction Fidelity.

There is an inherent trade-off between inverse problem solving and reconstruction fidelity. When the mask period is small, diffraction is strong: the encoding is rich and the spectral information is more explicitly separated, but the associated forward degradation is more severe, making the inverse problem harder to solve. Conversely, when the mask period is large, diffraction is weak: the encoding becomes insufficient, which simplifies the reconstruction numerically but reduces spectral fidelity. To quantify this trade-off, we compare ADIS reconstructions at $f=50mm$ for different mask periods. All models adopt the ODAUVST-3stg architecture and are trained on the CAVE-1024 and KAIST datasets.

As illustrated in Fig.~\ref{Fig4}, for $f=50 mm, s_{pixel}=3.45 \mu m$, mask periods in the  range $d=180-310 \mu m$ consistently yield high spectral reconstruction fidelity. Accordingly, in the main implementation we adopt a calculation-friendly and fabrication-friendly setting with a mask period of $d=200 \mu m$, which is used to build the physical prototype system.

Generalizing this observation, note that the effective diffraction strength of ADIS scales approximately with the ratio $f/d$. 
Therefore, for systems targeting the $450-650 nm$  range, we recommend choosing $f/d$ in the range of $161.29-277.78$, which empirically provides a good trade-off between sufficient spectral encoding and stable, accurate reconstruction.

\section{Snapshot Spectral Imaging with Aperture Diffraction}
\label{sec4}
\subsection{Forward Formation Model}
\label{sec4.1}
Our main objective is to capture Hyperspectral Images (HSIs) using a conventional RGB image sensor by placing a binary mask with orthogonal periodic distributions of amplitudes in aperture plane. Therefore, The forward formation model includes the response function through RGB mosaic filters, while the quantum-efficiency function for a monochromatic sensor can be used alternatively. Suppose that we want to capture HSIs ${x_{\lambda}}(m,n)$  from a Bayer-pattern measurement ${y_c}(m,n)$ with spectrally-varying point spread functions (PSFs) ${p_{\lambda}}(m,n)$. The sensor has the spectral sensitivity function ${\Omega _c}(\lambda )$ for each color channel $c \in \{ r,g,b\}$. Therefore, the forward formation model under normal exposure can be represented as:

\begin{equation}
\setlength{\abovedisplayskip}{0.1cm}
\setlength{\belowdisplayskip}{0.1cm}
    \label{4-1}
    y_c^N(m,n) = \int {{\Omega _c}(\lambda )} ({x_\lambda }(m,n)*{p_\lambda }(m,n))d\lambda
\end{equation}

\noindent
where $*$ is defined as the convolution operator.

In ADIS, we can write image forward formation model in a discrete vector-and-matrix form. We denote the captured Bayer-pattern measurement as $\mathbf{y} \in {\mathbb{R}^{HW\times 1}}$, where $H, W$ denote the measurement's height, width. Let $\mathbf{x} \in {\mathbb{R}^{HW\Lambda \times 1}}$ be the HSIs vector, where $\Lambda$ is the number of wavelength channels.  
We can represent the sensor spectral sensitivity ${\Omega _c}(\lambda )$ and the convolution by the PSFs ${p_{\lambda}}(m,n)$ as matrices $\mathbf{\Omega}  \in {\mathbb{R}^{HW \times HW\Lambda}}$ and $\mathbf{P} \in {\mathbb{R}^{HW\Lambda \times HW\Lambda}}$, respectively. Given the sensing matrix $\mathbf{\Phi_1} \in {\mathbb{R}^{HW \times HW\Lambda}}$ as the product of $\mathbf{\Omega}$ and $\mathbf{P}$ determined by the binary masks, the degradation model of ADIS can be formulated as:
\begin{equation}
\setlength{\abovedisplayskip}{0cm}
\setlength{\belowdisplayskip}{0cm}
    \label{4-2}
    \mathbf{y}=\mathbf{\Phi}\mathbf{x}+\mathbf{n}
\end{equation}
\noindent
where $\mathbf{n}$ represents the vectorized imaging noise on the measurement. Then the task of HSIs reconstruction is given $\mathbf{y}$ and $\mathbf{\Phi}$ (calculate based on hardware parameter), solving $\mathbf{x}$.

\begin{figure*}[t]
    \begin{center}
    \includegraphics[width=1\linewidth]{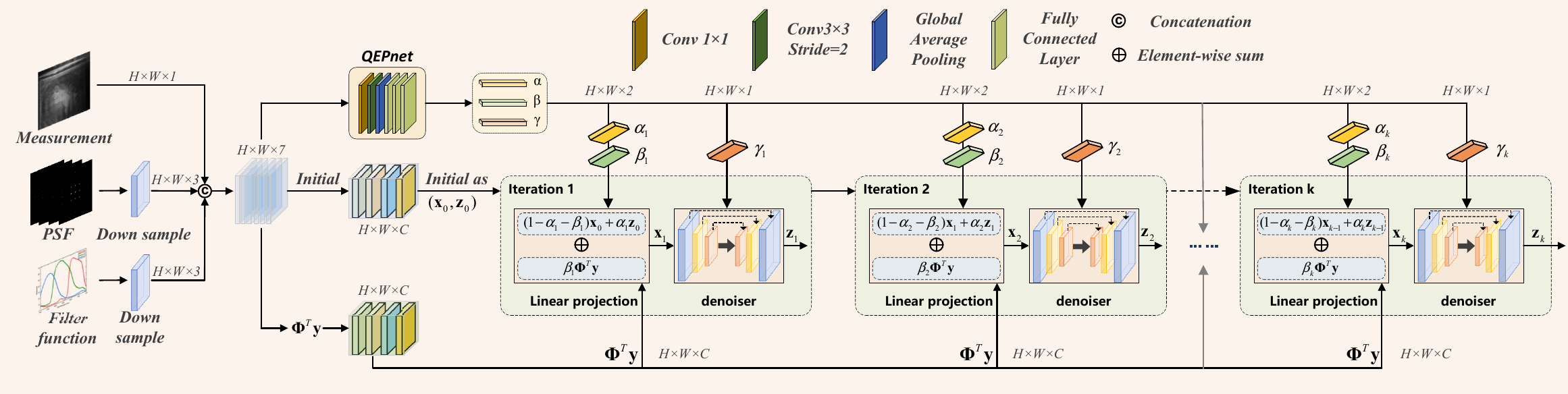}
    \end{center}
\vspace{-0.3cm} 
\caption{Illustration of ODAUF with k stages. Theoretically, the denoisers in ODAUF as learnable regularization terms can be replaced with different designs.}
\label{Fig5.1}
\vspace{-0.3cm} 
\end{figure*}

\subsection{Calibration-Free Reconstruction}
\label{sec4.1.2}

The diffraction of the ODL is jointly determined by the mask design and the employed lens configuration. While theoretically computed PSFs are usually derived under the ideal thin-lens assumption, practical lenses inevitably introduce wavelength-dependent convolutional degradations arising from diffraction, aberrations, and other optical imperfections. Given the specific convolution degradation ${\delta _\lambda ^{Lens}(m,n)}$ introduced by non-ideal lenses in each wavelength band, the actual forward process in a real system can be described as follows:

\begin{equation}
\setlength{\abovedisplayskip}{-0.2cm}
\setlength{\belowdisplayskip}{-0.2cm}
\label{eq1.1.2}
    {y_c^{{\mathop{\rm Re}\nolimits} al}(m,n) = \int {{\Omega _c}(\lambda )} ({x_\lambda }(m,n)*p_\lambda ^{{\mathop{\rm Re}\nolimits} al}(m,n))d\lambda }
\end{equation}

\begin{equation}
\setlength{\abovedisplayskip}{-0.2cm}
\setlength{\belowdisplayskip}{-0.2cm}
\label{eq1.1.3}
    {p_\lambda ^{{\mathop{\rm Re}\nolimits} al}(m,n) = p_\lambda ^{Simu}(m,n)*\delta _\lambda ^{Lens}(m,n)}
\end{equation}

Accordingly, for the network trained using theoretically simulated PSFs $p_\lambda ^{Simu}(m,n)$, the reconstruction process on real measurements can be interpreted as assuming that the observed measurements  $y_c^{{\mathop{\rm Re}\nolimits} al}(m,n)$ are generated by a forward model encoded by the simulated PSFs:

\begin{equation}
\setlength{\abovedisplayskip}{-0.2cm}
\setlength{\belowdisplayskip}{-0.2cm}
\label{eq1.1.4}
    \int {{\Omega _c}(\lambda )} \{ \underbrace {[{x_\lambda }(m,n)*\delta _\lambda ^{Lens}(m,n)]}_{Recon.~Objectives}*p_\lambda ^{Simu}(m,n)\} d\lambda
\end{equation}

\vspace{0.2cm}

In this case, the network tends to reconstruct ${x_\lambda }(m,n)*\delta _\lambda ^{Lens}(m,n)$. Consequently, the overall output can be interpreted as SSI under the finite imaging capability of the employed lens.

\subsection{Orthogonal Diffraction-Aware Unfolding Framework}
\label{sec4.2}

\subsubsection{\textbf{Optimization Analysis}}
To leverage the ADIS diffractive degradation patterns to adjust the iterative learning, we formulate a principled Orthogonal Diffraction-Aware Unfolding Framework (ODAUF) as shown in Fig.~\ref{Fig5.1}. In particular, the original HSIs signals could be estimated by minimizing the following energy function as:

\begin{equation}
\setlength{\abovedisplayskip}{0cm}
\setlength{\belowdisplayskip}{0cm}
    \hat {\mathbf{x} } = \mathop {argmin }\limits_{\mathbf{x}} \frac{1}{2}{\left\| {\mathbf{y} - {\bf{\Phi}} \mathbf{x}} \right\|^2} + \tau R(\mathbf{x})
\label{4c-1}
\end{equation}

\noindent 
where $\frac{1}{2}{\left\| {\mathbf{y} - {\bf{\Phi}} \mathbf{x}} \right\|^2}$ represents the data fidelity error of the forward imaging model, $R({\mathbf{x}})$ indicates the regularization terms, $\tau$ is hyperparameter balancing the importance. Here, $R({\mathbf{x}})$ represents unknown prior functions that can be sparsity, TV, low-rank, and the network-based image prior. 
 
The reconstruction of HSIs $\mathbf{x}$ can be effectively addressed via the Half-Quadratic Splitting (HQS) method
By introducing auxiliary variables $\mathbf{z}$ to decompose the problem, Equation \ref{4c-1} can be reformulated as:

\begin{equation}
\setlength{\abovedisplayskip}{0cm}
\setlength{\belowdisplayskip}{0cm}
\begin{aligned}
    \hat {\mathbf{x} }  = \mathop {argmin }\limits_{\mathbf{x}, \mathbf{z}} \frac{1}{2}{\left\| {\mathbf{y} - {\bf{\Phi }} \mathbf{x}} \right\|^2} + \tau R(\bf{z}), \text{ \bf{s.t. }} \bf{z} = \bf{x}
\end{aligned}
\label{4c-2}
\end{equation}

This is a constrained optimization problem. Here, we adopt the half-quadratic splitting (HQS) algorithm for its simplicity and fast convergence to obtain an unfolding inference. By introducing an auxiliary variable $\bf{z}$, Equation~\ref{4c-2} is solved by minimizing:

\begin{equation}
\setlength{\abovedisplayskip}{0cm}
\setlength{\belowdisplayskip}{0cm}
\begin{aligned}
    (\hat {\bf{x}},\hat {\bf{z}}) = \mathop {argmin{\rm{ }}}\limits_{{\bf{x}},{\bf{z}}} \frac{1}{2}{\left\| {{\bf{y}} - {\bf{\Phi x}}} \right\|^2} + \tau R({\bf{z}}) + \frac{\rho }{2}{\left\| {{\bf{z}} - {\bf{x}}} \right\|^2}
\end{aligned}
\label{4c-2.5}
\end{equation}

\noindent
where $\rho$ is a penalty parameter that forces $\bf{x}$ and $\bf{z}$ to approach the fixed point. Such a problem can be addressed by iteratively solving subproblems for $\bf{x}$ and $\bf{z}$:

\begin{equation}
\setlength{\abovedisplayskip}{0cm}
\setlength{\belowdisplayskip}{0cm}
    {{\bf{x}}^{k + 1}} = \mathop {argmin}\limits_{\bf{x}} {\left\| {{\bf{y}} - {\bf{\Phi x}}} \right\|^2} + \rho {\left\| {{\bf{x}} - {{\bf{z}}^k}} \right\|^2}
\label{4c-3}
\end{equation}

\begin{equation}
\setlength{\abovedisplayskip}{0cm}
\setlength{\belowdisplayskip}{0cm}
    {{\bf{z}}^{k + 1}} = \mathop {argmin{\rm{ }}}\limits_{\bf{z}}  \frac{\rho }{2}{\left\| {{\bf{z}} - {{\bf{x}}^{k + 1}}} \right\|^2} + \tau R({\bf{z}})
\label{4c-4}
\end{equation}
\noindent
where $k = 0,1,2,...,K - 1$ indexes the iteration. According to Equation~\ref{4c-3}, $\rho$ should be large enough so that $\bf{x}$ and $\bf{z}$ are approximately equal to the fixed point. However, this would also result in slow convergence. Therefore, a good rule of thumb is to iteratively increase $\rho$. 

\subsubsection{\textbf{ODAUF}}
Note that the data fidelity term is associated with a quadratically regularized least-squares problem, i.e., $\mathbf{x}_{k+1}$ in Equation \ref{4c-3}. It has a closed-form solution as:

\begin{equation}
\setlength{\abovedisplayskip}{0cm}
\setlength{\belowdisplayskip}{0cm}
    {{\bf{x}}^{k + 1}} = {({\bf{\Phi}}^T{\bf{\Phi } } + \rho \mathbf{I} )^{ - 1}}({\bf{\Phi }}^T{\mathbf{y}} + \rho {{\bf{z}}^k})
\label{4c-5}
\end{equation}

\noindent
where ${( {\bf{\Phi}}^T{\bf{\Phi}} + \rho I)} \in {\mathbb{R}^{HW\Lambda  \times HW\Lambda}}$, $({\bf{\Phi}}^T{\bf{\Phi}} + \rho {{\bf{z}}^k}) \in {R^{HW\Lambda  \times 1}}$. Directly using the closed-form solution in Equation~\ref{4c-5} for iteration is computationally prohibitive. Instead, the iterative conjugate gradient (CG) method accelerates convergence with a one-step gradient descent, achieving sufficient proximity to a local optimum~\cite{compact, dong2018denoising}. In this way, the solution of Equation~\ref{4c-3} can be expressed as:

\begin{equation}
\setlength{\abovedisplayskip}{0cm}
\setlength{\belowdisplayskip}{0cm}
    \begin{array}{c}
    {{\bf{x}}^{k + 1}} = {{\bf{x}}^k} - \delta [{{\bf{\Phi }}^T}({\bf{\Phi }}{{\bf{x}}^k} - {\bf{y}}) + \rho ({{\bf{x}}^k} - {{\bf{z}}^k})]\\
     = {\bf{\bar \Phi }}{{\bf{x}}^k} + \delta {{\bf{\Phi }}^T}{\bf{y}} + \delta \rho {{\bf{z}}^k}
\end{array}
\label{4c-6}
\end{equation}

\noindent
where ${\bf{\bar \Phi }} = [(1 - \delta \rho ){\bf{I}} - \delta {{\bf{\Phi }}^T}{\bf{\Phi }}] \in {\mathbb{R}^{HW\Lambda  \times HW\Lambda}}$ and $\delta$ is the gradient descent step size. Through the linear projection in Equation~\ref{4c-6}, ${{\bf{x}}^{k + 1}}$ can be updated very efficiently.

Additionally, setting $\tau$ as iteration-specific parameter as well, then the Equation~\ref{4c-4} can be reformulated as
\begin{equation}
\setlength{\abovedisplayskip}{0cm}
\setlength{\belowdisplayskip}{0cm}
    {{\bf{z}}^{k + 1}} = \mathop {argmin}\limits_{\bf{z}} \frac{1}{{2{{\left( {\sqrt {{\tau _{k + 1}}/{\rho _{k + 1}}} } \right)}^2}}}\left\| {{\bf{z}} - {{\bf{x}}^{k + 1}}} \right\|_2^2 + R({\bf{z}})
\label{4c-7}
\end{equation}

From the perspective of Bayesian probability, Equation~\ref{4c-7} can be treated as a denoising problem that image ${{\bf{z}}^{k + 1}}$ with a Gaussian noise at level ${\sqrt {{\tau _{k + 1}}/{\rho _{k + 1}}} }$. We set $\frac{1}{{{{\left( {\sqrt {{\tau _{k + 1}}/{\rho _{k + 1}}} } \right)}^2}}}$ as parameters to be estimated from ADIS. Let $\alpha  \buildrel def \over = [{\alpha _1},...,{\alpha _K}]$, ${\alpha _k} \buildrel def \over = {\rho _k}$, $\gamma \buildrel def \over = [{\gamma _1},...,{\gamma _K}]$, ${\gamma _k} \buildrel def \over = {\delta _k}{\rho _k}$, $\beta  \buildrel def \over = [{\beta _1},...,{\beta _K}]$ and ${\beta _k} \buildrel def \over = {\mu _k}/{\tau _k}$, then an approximation to the structure of the unfolding frame can be obtained: 

\begin{equation}
\setlength{\abovedisplayskip}{0.2cm}
\setlength{\belowdisplayskip}{0.2cm}
    \begin{array}{*{20}{c}}
    \left\{ {\begin{array}{*{20}{c}}
    {(\alpha ,\beta ,\gamma ) = \Theta ({\bf{y}},{\bf{\Phi }}),}\\
    {{{\bf{x}}^{k + 1}} = \Xi ({\bf{y}},{{\bf{z}}^k},{\alpha _{k + 1}},{\gamma _{k + 1}},{\bf{\Phi }}),}\\
    {{{\bf{z}}^{k + 1}} = {\cal D}({{\bf{x}}^{k + 1}},{\beta _{k + 1}}).}
\end{array}} \right.
    \end{array}
\label{4c-8}
\end{equation}
\noindent
where $\Theta$ denotes the parameter estimator that takes the compressed measurement $\bf{y}$ and the sensing matrix $\bf{\Phi}$ of the ADIS as inputs. $\bf{y}$ is the measurement after convolution with polychromatic PSFs and integration along temporal dimension. $\Xi$ equivalent to Equation \ref{4c-6} denotes the linear projection to conduct an update of $\bf{x}$, and $\mathcal{D}$ represents the Gaussian denoiser solving Equation \ref{4c-7}. Parameters $\alpha$ and $\gamma$ estimated by $\Theta$ direct the unfolding learning by adaptively scaling the linear projection in Equation~\ref{4c-6} and providing variable noise levels for the denoiser prior in Equation~\ref{4c-7}.

\begin{figure*}[t]
    \begin{center}
    \includegraphics[width=1\linewidth]{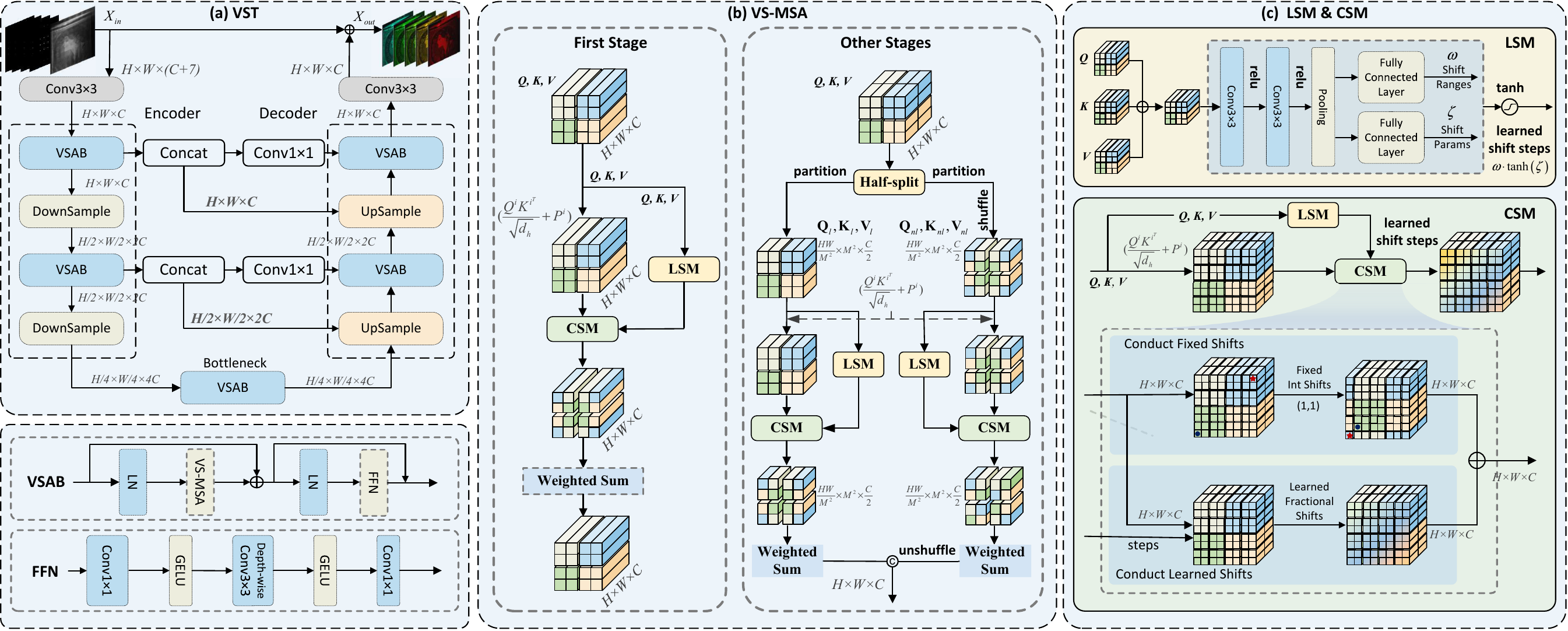}
    \end{center}
\vspace{-0.3cm} 
\caption{(a) Diagram of VST. VST adopts a three-layer U-shaped structure. VSAB consists of an FFN, a VS-MSA, and two layer normalizations. (b) VS-MSA employs parameters learned via LSM to perform feature shifting (CSM) and self-attention calculations on input features. (c) Illustration of LSM and CSM.}
\label{Fig5.2}
\vspace{-0.3cm} 
\end{figure*}

\subsection{Realization of ODAUF}
To address the challenges of solving the complex inverse problem caused by consistent diffractive aliasing, we develop an Orthogonal Diffraction-Aware Unfolding Framework (ODAUF) that comprehensively perceives the diffraction degradation process of ADIS for efficient solving. The ODAUF is illustrated in Fig. \ref{Fig5.1}.
First, inspired by the success of deep unfolding methods on denoising tasks~\cite{dong2018denoising} and computational spectral imaging (CSI) tasks~\cite{DAUHST}, we design a lightweight Hyperparameter Estimation Module (HEM) as $\Theta$ to extract key cues from the compressed measurements and priori information for subsequent iterations.
Notably, the computed PSFs exhibit greater spatial extent than the input filter function. To tackle data redundancy, we first downsample the PSF's spatial resolution and the filter function's channel dimension as $256\times256\times3$ to match the network body structure. 
Fig. \ref{Fig5.1} illustrates the architecture of HEM, which includes a $conv1 \times 1$, a $conv3 \times 3$, a global average pooling, and three fully connected layers. The estimated parameters $\alpha  \buildrel def \over = [{\alpha _1},...,{\alpha _K}]$, $\gamma \buildrel def \over = [{\gamma _1},...,{\gamma _K}]$, and $\beta  \buildrel def \over = [{\beta _1},...,{\beta _K}]$ are repeated as multichannel feature maps that have the same resolution as the input features, whose number of channel layers is kept consistent with the number of iterations, allowing the estimated parameters to guide and optimize the reconstruction process. 
The initial values for the iterative process in ODAUF are acquired through a multi-scale integration of measurements and prior knowledge. During the iterative learning process, the denoiser is embedded in the ODAUF to act as a regular term to fully utilize the guiding role of $\beta$.

\subsection{Voxel Shift Transformer}
The utilization of transformer models for global and local perception encounters challenges of a restricted receptive field and computationally intensive processing.
The denoiser plays a pivotal role in enhancing feature extraction and expanding the receptive field while maintaining the efficiency of ODAUF. Previously, we introduced the Shift-Shuffle Transformer (SST)~\cite{ADIS} as a denoiser, employing fixed integer shifts to leverage the orthogonal diffraction patterns of ADIS for improved reconstruction. However, SST lacks adaptability in adjusting shift strategies based on input features, necessitating a higher number of iterations to achieve optimal performance.

To address these challenges, we introduce the Voxel Shift Transformer (VST), an adaptive degradation-aware denoiser embedded in ODAUF. By leveraging voxel shifts, VST efficiently extracts diffraction degradation features and high-dimensional information, reducing iteration counts while significantly lowering computational and memory costs.
Structurally, it adopts a three-layer U-shape design, similar to \cite{MST,DAUHST,ADIS}, as illustrated in Fig.~\ref{Fig5.2}(a). VST leverages channel shuffling along with integer and fractional shift operations to implement an attention mechanism for 3D voxel fractional shift enhancement, improving both global feature extraction and long- and short-range modeling. While learning shift step sizes at each attention computation adds computational and memory costs, the enhanced denoising capability at each level enables ODAUVST-5stg to surpass CSST-9stg in reconstruction performance, ultimately reducing overall computational and storage demands.

First, to incorporate hardware priors and imaging process priors into the denoiser and HEM, VST undergoes an initialization process (which is performed for all comparison algorithms to equally benefit from additional priors). Specifically, the $\varsigma \in {\mathbb{R}^{4H \times 4W \times C}}$ are downsampled via $conv 4\times 4$ as shape $H \times W \times 3$. Similarly, the filter function $\varsigma \in {\mathbb{R}^{H \times W \times C}}$ are downsampled via $conv 2\times 2$ as shape $H \times W \times 3$. These are then concatenated with the measurement $y \in {\mathbb{R}^{H \times W \times 1}}$, followed by an upsampling operation to form $X_0 \in {\mathbb{R}^{H \times W \times C}}$.
Secondly, ${X_0}$ passes through the encoder, bottleneck, and decoder to be embedded into deep features$X_f \in {\mathbb{R}^{H \times W \times C}}$. The basic unit Voxel Shift Attention Block (VSAB) assumes the composition of encoder and decoder.

\subsubsection{\textbf{Initialization of Voxel Shift Attention Block}}
VSAB consists of two layer normalizations (LN), a VS-MSA, and a Feed-Forward Network (FFN) following the classic design. The most important part of VSAB is Voxel Shift Multi-head Self-Attention(VS-MSA) with two modes on  first stage and other stage as shown in Fig.~\ref{Fig5.2}(b). The input tokens of VS-MSA are denoted as ${X_{in}} \in {\mathbb{R}^{H \times W \times {\rm{C}}}}$. 
Similar to \cite{ADIS, DAUHST}, ${X_{in}}$ is subsequently linearly projected into query ${Q \in {\mathbb{R}^{H \times W \times {\rm{C}}}}}$, key ${K \in {\mathbb{R}^{H \times W \times {\rm{C}}}}}$, and value ${V \in {\mathbb{R}^{H \times W \times {\rm{C}}}}}$ as:

\begin{equation}
\setlength{\abovedisplayskip}{0cm}
\setlength{\belowdisplayskip}{0cm}
    Q = {X_{in}}{W^Q},K = {X_{in}}{W^K},V = {X_{in}}{W^V}.
\label{1-10}
\end{equation}

\noindent
where ${W^Q},{W^K},{W^V} \in {\mathbb{R}^{C \times C}}$ are learnable parameters and biases are omitted for simplification.

\subsubsection{\textbf{First Stage}}
In the first stage of VST, only shift operations are performed in the latter two dimensions without shuffle operations. 
The input tokens ${X_{in}} \in {\mathbb{R}^{H \times W \times {\rm{C}}}}$ are first linearly projected as ${Q_1},{K_1},{V_1}$, and then VSAB learns the fractional shift parameters ${\upsilon _1}$ by feeding ${Q_1},{K_1},{V_1}$ into the learning shifts module (LSM). Then the self-attention of first stage $A_1^i$ is calculated inside each head as:

\begin{equation}
\setlength{\abovedisplayskip}{-0.1cm}
\setlength{\belowdisplayskip}{-0.1cm}
\label{1-11}
    A_1^i = {\rm{softmax}}(\Upsilon (\frac{{Q_1^iK_1^{{i^T}}}}{{\sqrt {{d_h}} }} + P_1^i,{\upsilon _1}))V_1^i
\end{equation}

\noindent
Where ${\upsilon _1} = {\cal L}({Q_1},{K_1},{V_1})$, $h=1$, $d_h=C$, $i = 1,2,...,h$, ${\cal L}$ denotes the LSM, $\Upsilon \left( {X,\upsilon } \right)$ denotes conduct shifts module (CSM) that shifts the input feature map $X$ according to the learned shift parameter $\upsilon $. Additionally, the first stage computes MSA with position-specific windows of size $M \times M$, $P_1^i \in {\mathbb{R}^{{M^2} \times {M^2}}}$ are learnable position parameters. Then the output of first stage is $S{({X_{in}})_1} = \sum\limits_{i = 1}^h {A_1^iW_1^i}$.

\subsubsection{\textbf{Other Stage}}
Q, K, V will be split into two equal parts along the channel dimension as: ${Q_o} = [{Q_{of}},{Q_{os}}],{K_o} = [{K_{of}},{K_{os}}],{V_o} = [{V_{of}},{V_{os}}]$. For ${Q_{of}},{K_{of}},{V_{of}} \in {\mathbb{R}^{H \times W \times \frac{C}{2}}}$, we first partition them into non-overlapping windows of size $M \times M$ and reshape them as ${\mathbb{R}^{\frac{{HW}}{{{M^2}}} \times {M^2} \times \frac{C}{2}}}$. Then the self-attention calculated by ${Q_{of}},{K_{of}},{V_{of}}$ inside each head is:
\begin{equation}
\setlength{\abovedisplayskip}{0cm}
\setlength{\belowdisplayskip}{0cm}
\label{1-12}
    A_{of}^i \!=\! {\rm{softmax}}(\Upsilon (\frac{{Q_{of}^iK_{of}^{{i^T}}}}{{\sqrt {{d_h}} }} \!+\! P_{of}^i,{\upsilon _{of}}))V_{of}^i
\end{equation}
\noindent
where ${\upsilon _{of}} = {\cal L}({Q_{of}},{K_{of}},{V_{of}})$, ${d_h} = \frac{C}{{2h}}$, $i = 1,2,...,h$, $h = 1$ for second stage and $h = 2$ for bottleneck. Similarly, $P_o^i \in {R^{{M^2} \times {M^2}}}$ are learnable position parameters.

for ${Q_{os}},{K_{os}},{V_{os}} \in {R^{H \times W \times \frac{C}{2}}}$, we compute cross-window interactions and richer shift enhancements by shuffle operations inspired by ShuffleNet and DAUHST~\cite{DAUHST}. Particularly, ${Q_{os}},{K_{os}},{V_{os}}$ are partitioned as ${R^{\frac{{HW}}{{{M^2}}} \times {M^2} \times \frac{C}{2}}}$ and then shuffled as ${R^{{M^2} \times \frac{{HW}}{{{M^2}}} \times \frac{C}{2}}}$ to establish inter-window dependencies. Then the self-attention calculated by ${Q_{os}},{K_{os}},{V_{os}}$ is:

\begin{equation}
\setlength{\abovedisplayskip}{0cm}
\setlength{\belowdisplayskip}{0cm}
\label{1-13}
    A_{os}^i \!\!=\!\! {\Lambda ^T}\!({\rm{softmax}}(\Upsilon (\frac{{\Lambda (Q_{os}^i)\Lambda (K_{os}^{{i^T}})}}{{\sqrt {{d_h}} }} \!+\! P_{os}^i,{\upsilon _{os}}))\Lambda (V_{os}^i))
\end{equation}

\noindent
Where ${\upsilon _{os}} = {\cal L}({Q_{os}},{K_{os}},{V_{os}})$, $\Lambda \left(  \cdot  \right)$ denotes the channel shuffle operations. And the overall self-attention is:
\begin{equation}
\setlength{\abovedisplayskip}{0cm}
\setlength{\belowdisplayskip}{0cm}
\label{1-14}
 S{({X_{in}})_o} = \sum\limits_{i = 1}^h {A_{of}^iW_{of}^i}  + \sum\limits_{i = 1}^h {A_{os}^iW_{os}^i}   
\end{equation}

Then we reshape the result of Equation \ref{1-14} to obtain the output ${X_{out}} \in {\mathbb{R}^{H \times W \times {\rm{C}}}}$. 

\subsubsection{\textbf{LSM and CSM}} 
The LSM extracts fractional shift parameters from input features, while the CSM enhances self-attention through fixed integer shifts $\mathcal{S}$ and learnable fractional shifts $\varsigma$ (Fig.~\ref{Fig5.2}(c)). Specifically, LSM captures local features using two $Conv 3\times3$, a global average pooling, and regresses shift parameters $\zeta$ through a fully connected layer, regresses shift ranges $\omega$ through another fully connected layer, with $tanh$ applied to calculate the learned fractional shift steps $\varsigma$: 
 
 \begin{equation}
\setlength{\abovedisplayskip}{0cm}
\setlength{\belowdisplayskip}{0cm}
\label{1-15}
   \varsigma = \omega  \cdot \tanh \left( \zeta  \right)
\end{equation}

 Then the fractional shifts in CSM are achieved through translational interpolation using $grid\_sample$ with normalized mesh mapping. 
 The CSM combines a fixed integer shift and a continuous, grid-based shift in a fully differentiable manner.
 In $grid \_ sample$, we adopt bilinear interpolation with reflection padding to avoid boundary artifacts and to preserve continuity at the image borders. The output of this grid-based warping is finally reshaped back to the input shape and added to the fixed-shift tensor, resulting in the final shift-enhanced self-attention.

\begin{table*}[t]
\caption{Quantitative comparison of reconstruction results of different algorithms, PSNR (dB) and SSIM are reported.}
\begin{center}
\renewcommand{\arraystretch}{1.1}
\resizebox{180mm}{32mm}{
\begin{tabular}{c|cccccccccccccc}
\hline
\rowcolor[HTML]{EFEFEF} 
\textbf{Algorithm}                                                    & \textbf{Inference Time}                        & \textbf{Params}                                  & \textbf{GFLOPS}                                                        & \textbf{S1}                           & \textbf{S2}                           & \textbf{S3}                           & \textbf{S4}                           & \cellcolor[HTML]{EFEFEF}\textbf{S5}   & \cellcolor[HTML]{EFEFEF}\textbf{S6}   & \cellcolor[HTML]{EFEFEF}\textbf{S7}   & \cellcolor[HTML]{EFEFEF}\textbf{S8}   & \cellcolor[HTML]{EFEFEF}\textbf{S9}   & \cellcolor[HTML]{EFEFEF}\textbf{S10}  & \cellcolor[HTML]{EFEFEF}\textbf{Avg}   \\ \hline
\rowcolor[HTML]{FFFFFF} 
\cellcolor[HTML]{EFEFEF}                                              & \cellcolor[HTML]{FFFFFF}                       & \cellcolor[HTML]{FFFFFF}                         & \cellcolor[HTML]{FFFFFF}                                               & 31.43                                 & 29.64                                 & 25.44                                 & 33.50                                 & 28.77                                 & 28.38                                 & 26.78                                 & 31.72                                 & {\color[HTML]{333333} 36.40}          & 27.96                                 & \textbf{30.00}                         \\
\rowcolor[HTML]{FFFFFF} 
\multirow{-2}{*}{\cellcolor[HTML]{EFEFEF}\textbf{UNet}~\cite{Unet}}               & \multirow{-2}{*}{\cellcolor[HTML]{FFFFFF}1ms}  & \multirow{-2}{*}{\cellcolor[HTML]{FFFFFF}23.32M} & \multirow{-2}{*}{\cellcolor[HTML]{FFFFFF}5.33}                         & {\color[HTML]{333333} 0.940}          & 0.912                                 & 0.877                                 & 0.914                                 & 0.897                                 & 0.884                                 & 0.852                                 & 0.916                                 & 0.936                                 & 0.892                                 & \textbf{0.902}                         \\ \hline
\rowcolor[HTML]{FFFFFF} 
\cellcolor[HTML]{EFEFEF}                                              & \cellcolor[HTML]{FFFFFF}                       & \cellcolor[HTML]{FFFFFF}                         & \cellcolor[HTML]{FFFFFF}{\color[HTML]{333333} }                        & 29.27                                 & 28.07                                 & {\color[HTML]{333333} 23.02}          & 31.53                                 & 27.40                                 & 26.29                                 & 23.52                                 & 30.06                                 & {\color[HTML]{333333} 32.67}          & 26.17                                 & \textbf{27.80}                         \\
\rowcolor[HTML]{FFFFFF} 
\multirow{-2}{*}{\cellcolor[HTML]{EFEFEF}\textbf{MIRNet}~\cite{MIRNet}}             & \multirow{-2}{*}{\cellcolor[HTML]{FFFFFF}1ms}  & \multirow{-2}{*}{\cellcolor[HTML]{FFFFFF}2.05M}  & \multirow{-2}{*}{\cellcolor[HTML]{FFFFFF}{\color[HTML]{333333} 14.66}} & {\color[HTML]{333333} 0.910}          & 0.870                                 & 0.842                                 & 0.898                                 & 0.873                                 & 0.858                                 & 0.833                                 & 0.905                                 & 0.854                                 & 0.844                                 & \textbf{0.869}                         \\ \hline
\rowcolor[HTML]{FFFFFF} 
\cellcolor[HTML]{EFEFEF}                                              & \cellcolor[HTML]{FFFFFF}                       & \cellcolor[HTML]{FFFFFF}                         & \cellcolor[HTML]{FFFFFF}{\color[HTML]{333333} }                        & 32.18                                 & {\color[HTML]{333333} 29.90}          & 26.30                                 & {\color[HTML]{333333} 34.20}          & 30.26                                 & 29.00                                 & 27.73                                 & 32.32                                 & 34.18                                 & 27.15                                 & \textbf{30.32}                         \\
\rowcolor[HTML]{FFFFFF} 
\multirow{-2}{*}{\cellcolor[HTML]{EFEFEF}\textbf{Lambda-Net}~\cite{l-net}}         & \multirow{-2}{*}{\cellcolor[HTML]{FFFFFF}1ms}  & \multirow{-2}{*}{\cellcolor[HTML]{FFFFFF}32.73M} & \multirow{-2}{*}{\cellcolor[HTML]{FFFFFF}{\color[HTML]{333333} 23.08}} & {\color[HTML]{333333} 0.927}          & 0.909                                 & 0.871                                 & 0.886                                 & 0.902                                 & 0.880                                 & 0.844                                 & 0.916                                 & 0.900                                 & 0.883                                 & \textbf{0.892}                         \\ \hline
\rowcolor[HTML]{FFFFFF} 
\cellcolor[HTML]{EFEFEF}                                              & \cellcolor[HTML]{FFFFFF}                       & \cellcolor[HTML]{FFFFFF}                         & \cellcolor[HTML]{FFFFFF}{\color[HTML]{333333} }                        & 32.36                                 & 29.94                                 & {\color[HTML]{333333} 25.63}          & 32.95                                 & 30.05                                 & 28.38                                 & 25.81                                 & 32.46                                 & {\color[HTML]{333333} 35.53}          & 28.87                                 & \cellcolor[HTML]{FFFFFF}\textbf{30.20} \\
\rowcolor[HTML]{FFFFFF} 
\multirow{-2}{*}{\cellcolor[HTML]{EFEFEF}\textbf{TSA-Net}~\cite{TSA-net}}            & \multirow{-2}{*}{\cellcolor[HTML]{FFFFFF}5ms}  & \multirow{-2}{*}{\cellcolor[HTML]{FFFFFF}44.25M} & \multirow{-2}{*}{\cellcolor[HTML]{FFFFFF}{\color[HTML]{333333} 92.00}} & {\color[HTML]{333333} 0.950}          & 0.916                                 & 0.884                                 & 0.917                                 & 0.916                                 & 0.887                                 & 0.858                                 & 0.928                                 & 0.940                                 & 0.905                                 & \cellcolor[HTML]{FFFFFF}\textbf{0.910} \\ \hline
\rowcolor[HTML]{FFFFFF} 
\cellcolor[HTML]{EFEFEF}                                              & \cellcolor[HTML]{FFFFFF}                       & \cellcolor[HTML]{FFFFFF}                         & \cellcolor[HTML]{FFFFFF}{\color[HTML]{333333} }                        & 32.81                                 & {\color[HTML]{333333} 30.00}          & 25.77                                 & 33.60                                 & 29.06                                 & 28.89                                 & 25.51                                 & 32.25                                 & {\color[HTML]{333333} 34.99}          & 29.34                                 & \textbf{30.22}                         \\
\rowcolor[HTML]{FFFFFF} 
\multirow{-2}{*}{\cellcolor[HTML]{EFEFEF}\textbf{MPRNet}~\cite{MPRNet}}             & \multirow{-2}{*}{\cellcolor[HTML]{FFFFFF}14ms} & \multirow{-2}{*}{\cellcolor[HTML]{FFFFFF}2.96M}  & \multirow{-2}{*}{\cellcolor[HTML]{FFFFFF}{\color[HTML]{333333} 77.70}} & {\color[HTML]{333333} 0.956}          & 0.920                                 & 0.901                                 & 0.917                                 & 0.908                                 & 0.913                                 & 0.879                                 & 0.938                                 & 0.922                                 & 0.916                                 & \textbf{0.917}                         \\ \hline
\rowcolor[HTML]{FFFFFF} 
\cellcolor[HTML]{EFEFEF}                                              & \cellcolor[HTML]{FFFFFF}                       & \cellcolor[HTML]{FFFFFF}                         & \cellcolor[HTML]{FFFFFF}{\color[HTML]{333333} }                        & 32.68                                 & 30.23                                 & {\color[HTML]{333333} 25.57}          & 31.21                                 & 29.84                                 & 28.71                                 & 26.58                                 & 32.52                                 & {\color[HTML]{333333} 36.03}          & 29.10                                 & \textbf{30.25}                         \\
\rowcolor[HTML]{FFFFFF} 
\multirow{-2}{*}{\cellcolor[HTML]{EFEFEF}\textbf{MST++}~\cite{MST++}}              & \multirow{-2}{*}{\cellcolor[HTML]{FFFFFF}2ms}  & \multirow{-2}{*}{\cellcolor[HTML]{FFFFFF}1.33M}  & \multirow{-2}{*}{\cellcolor[HTML]{FFFFFF}{\color[HTML]{333333} 17.52}} & {\color[HTML]{333333} 0.953}          & 0.923                                 & 0.902                                 & 0.909                                 & 0.912                                 & 0.914                                 & 0.881                                 & 0.941                                 & 0.928                                 & 0.929                                 & \textbf{0.919}                         \\ \hline
\rowcolor[HTML]{FFFFFF} 
\cellcolor[HTML]{EFEFEF}                                              & \cellcolor[HTML]{FFFFFF}                       & \cellcolor[HTML]{FFFFFF}                         & \cellcolor[HTML]{FFFFFF}{\color[HTML]{333333} }                        & {\color[HTML]{CB0000} \textbf{34.92}} & {\color[HTML]{333333} 31.57}          & 27.32                                 & 34.63                                 & {\color[HTML]{3531FF} \textbf{31.00}} & 30.20                                 & {\color[HTML]{000000} 27.94}          & 33.95                                 & 37.34                                 & 30.61                                 & \textbf{31.95}                         \\
\rowcolor[HTML]{FFFFFF} 
\multirow{-2}{*}{\cellcolor[HTML]{EFEFEF}\textbf{Restormer}~\cite{restormer}}          & \multirow{-2}{*}{\cellcolor[HTML]{FFFFFF}10ms} & \multirow{-2}{*}{\cellcolor[HTML]{FFFFFF}15.12M} & \multirow{-2}{*}{\cellcolor[HTML]{FFFFFF}{\color[HTML]{333333} 87.93}} & {\color[HTML]{CB0000} \textbf{0.967}} & 0.936                                 & 0.927                                 & 0.931                                 & {\color[HTML]{3531FF} \textbf{0.931}} & 0.933                                 & {\color[HTML]{3531FF} \textbf{0.896}} & {\color[HTML]{333333} 0.948}          & 0.955                                 & 0.949                                 & \textbf{0.937}                         \\ \hline
\rowcolor[HTML]{FFFFFF} 
\cellcolor[HTML]{EFEFEF}                                              & \cellcolor[HTML]{FFFFFF}                       & \cellcolor[HTML]{FFFFFF}                         & \cellcolor[HTML]{FFFFFF}{\color[HTML]{333333} }                        & {\color[HTML]{3531FF} \textbf{33.80}} & {\color[HTML]{3531FF} \textbf{33.70}} & {\color[HTML]{3531FF} \textbf{28.74}} & {\color[HTML]{3531FF} \textbf{34.77}} & {\color[HTML]{000000} 30.19}          & {\color[HTML]{3531FF} \textbf{31.24}} & {\color[HTML]{CB0000} \textbf{28.29}} & {\color[HTML]{3531FF} \textbf{34.87}} & {\color[HTML]{3531FF} \textbf{37.96}} & {\color[HTML]{3531FF} \textbf{31.91}} & {\color[HTML]{3531FF} \textbf{32.55}}  \\
\rowcolor[HTML]{FFFFFF} 
\multirow{-2}{*}{\cellcolor[HTML]{EFEFEF}\textbf{CSST-9stg(ICCV23)}~\cite{ADIS}}  & \multirow{-2}{*}{\cellcolor[HTML]{FFFFFF}17ms} & \multirow{-2}{*}{\cellcolor[HTML]{FFFFFF}6.56M}  & \multirow{-2}{*}{\cellcolor[HTML]{FFFFFF}{\color[HTML]{333333} 70.50}} & {\color[HTML]{3531FF} \textbf{0.966}} & {\color[HTML]{3531FF} \textbf{0.965}} & {\color[HTML]{3531FF} \textbf{0.946}} & {\color[HTML]{3531FF} \textbf{0.933}} & 0.927                                 & {\color[HTML]{3531FF} \textbf{0.945}} & 0.893                                 & {\color[HTML]{3531FF} \textbf{0.959}} & {\color[HTML]{3531FF} \textbf{0.957}} & {\color[HTML]{3531FF} \textbf{0.958}} & {\color[HTML]{3531FF} \textbf{0.945}}  \\ \hline
\rowcolor[HTML]{EBE5DD} 
\cellcolor[HTML]{EBE5DD}                                              & \cellcolor[HTML]{EBE5DD}                       & \cellcolor[HTML]{EBE5DD}                         & \cellcolor[HTML]{EBE5DD}                                               & {\color[HTML]{000000} 33.71}          & {\color[HTML]{CB0000} \textbf{34.77}} & {\color[HTML]{CB0000} \textbf{29.51}} & {\color[HTML]{CB0000} \textbf{36.36}} & {\color[HTML]{CB0000} \textbf{32.45}} & {\color[HTML]{CB0000} \textbf{31.95}} & {\color[HTML]{3531FF} \textbf{27.95}} & {\color[HTML]{CB0000} \textbf{35.09}} & {\color[HTML]{CB0000} \textbf{38.51}} & {\color[HTML]{CB0000} \textbf{32.85}} & {\color[HTML]{CB0000} \textbf{33.31}}  \\
\rowcolor[HTML]{EBE5DD} 
\multirow{-2}{*}{\cellcolor[HTML]{EBE5DD}\textbf{ODAUVST-5stg(Ours)}} & \multirow{-2}{*}{\cellcolor[HTML]{EBE5DD}17ms} & \multirow{-2}{*}{\cellcolor[HTML]{EBE5DD}4.53M}  & \multirow{-2}{*}{\cellcolor[HTML]{EBE5DD}53.64}                        & 0.964                                 & {\color[HTML]{CB0000} \textbf{0.971}} & {\color[HTML]{CB0000} \textbf{0.957}} & {\color[HTML]{CB0000} \textbf{0.943}} & {\color[HTML]{CB0000} \textbf{0.937}} & {\color[HTML]{CB0000} \textbf{0.951}} & {\color[HTML]{CB0000} \textbf{0.900}} & {\color[HTML]{CB0000} \textbf{0.959}} & {\color[HTML]{CB0000} \textbf{0.961}} & {\color[HTML]{CB0000} \textbf{0.968}} & {\color[HTML]{CB0000} \textbf{0.951}}  \\ \hline
\end{tabular}
}
\end{center}
\label{tab1}
\vspace{-0.4cm} 
\setlength{\belowcaptionskip}{-0.4cm}
\end{table*}

\begin{figure*}[t]
    \begin{center}
    \includegraphics[width=1\linewidth]{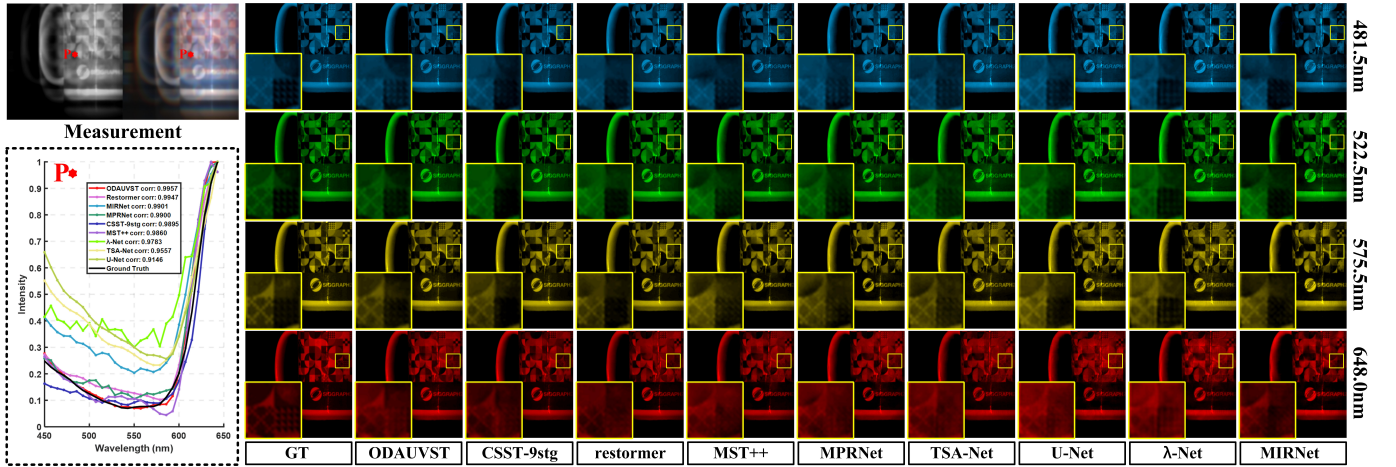}
    \end{center}
\vspace{-0.3cm} 
\caption{Simulation HSI reconstruction comparisons of different scenes with 10 (out of 28) spectral channels on ADIS. The left side displays the RGB image of the scene alongside the simulated measurements, while the central side presents a comparison between the reconstructed intensity-spectral curves at marked points on the RGB image and the corresponding Ground Truth.}
\label{Fig6}
\vspace{-0.1cm} 
\end{figure*}

\vspace{-0.2cm}
\section{Simulation Experiment Analysis}
\label{sec6}
To demonstrate the superiority of ADIS and the effectiveness of ODAUVST in solving the inverse problems of SSI, we conduct three key comparisons: (1) comparison of ODAUVST with other state-of-the-art (SOTA) algorithms, (2) comparison of ADIS with other representative compact SSI systems, and (3) comparison of various deep unfolding frameworks with ODAUF. Similar to previous works ~\cite{TSA-net, MST, DAUHST}, we select 28 wavelengths ranging from 450nm to 650nm and derive them via spectral interpolation for HSIs. To maintain consistent experimental conditions, we use PSFs of size $1024 \times 1024\times 28$ and HSI of size $512 \times 512\times 28$ for convolutional projection.  
From this projection, we extract the central $256\times 256 \times 28$ region, apply a spectral filter of matching size ($256\times 256 \times 28$), and integrate along the spectral dimension to produce a two-dimensional ($256\times 256$) input for the reconstruction algorithms. This image formation model maintains consistency while preparing rendered measurements for reconstruction.
We adopt two datasets, i.e., CAVE-1024~\cite{TSA-net} and KAIST~\cite{KAIST} for simulation experiments. 10 scenes from the KAIST dataset are selected for testing, while the CAVE-1024 dataset and another 20 scenes from the KAIST dataset are selected for training. We implement ODAUVST by Pytorch. All ODAUVST models are trained with Adam~\cite{ADAM} optimizer (${\beta _1} = 0.9$ and ${\beta _2} = 0.999$) using the Cosine Annealing scheme~\cite{sgdr} for 300 epochs on an RTX 4090D GPU. The initial learning rate is $8 \times {10^{ - 4}}$.

\vspace{-0.2cm}
\subsection{Comparison with Other Reconstructions}
\label{sec6.1}
As shown in Table~\ref{tab1}, we compare the results of ODAUVST with 8 methods including four reconstruction methods (lambda-Net~\cite{l-net}, TSA-Net~\cite{TSA-net}, MST++~\cite{MST++}, CSST~\cite{ADIS}), four Super-resolution algorithms (Unet~\cite{Unet}, Restormer~\cite{restormer}, MPRNet~\cite{MPRNet}, MIRNet\cite{MIRNet}) across 10 simulated scenes derived from selected KAIST dataset HSIs.

\begin{figure*}[t]
    \begin{center}
    \includegraphics[width=1\linewidth]{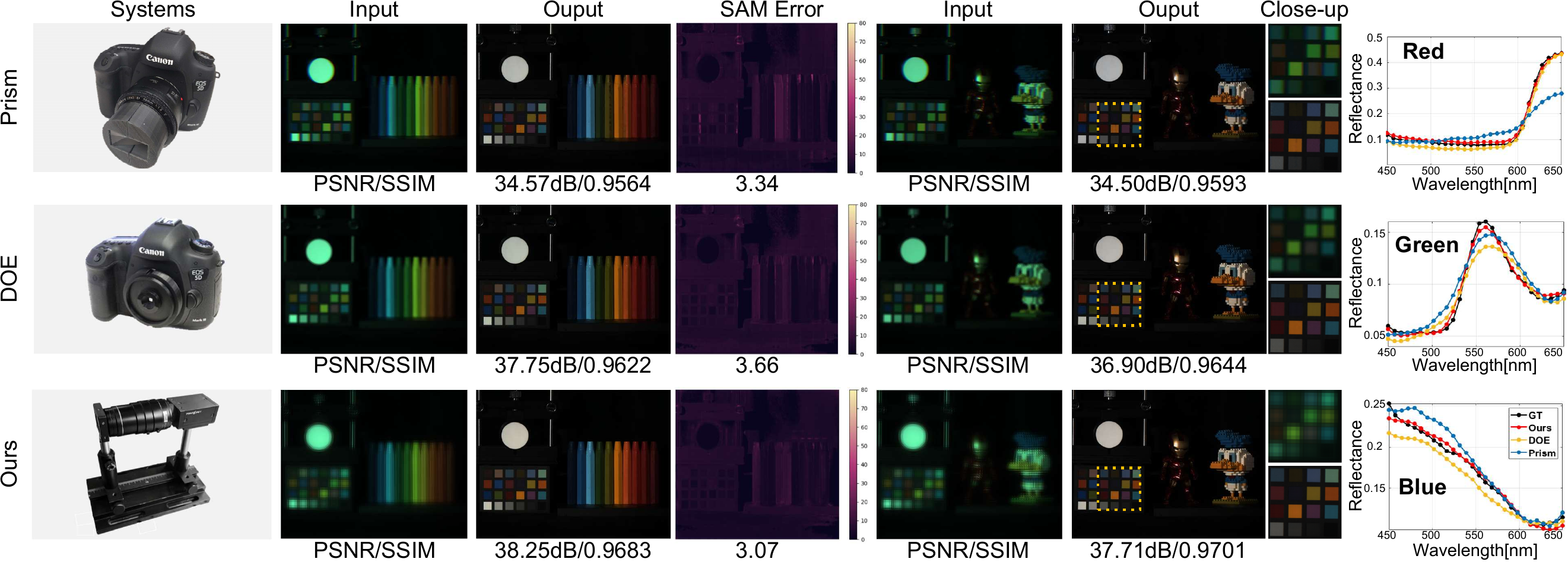}
    \end{center}
\vspace{-0.3cm} 
\caption{Comparison of ADIS and two different existing compact computational spectral imaging methods, a DOE-based system and a prism-based system with ten ground-truth spectral images.}
\label{Fig7}
\vspace{-0.4cm} 
\setlength{\belowcaptionskip}{-0.4cm}
\end{figure*}

\subsubsection{Quantitative Comparisons}
\label{sec6.1.1}

As shown in Table~\ref{tab1}, ODAUVST demonstrates the best performance on the ADIS spectral reconstruction, i.e., 33.31dB in PSNR and 0.951 in SSIM. ODAUVST-5stg significantly outperforms two recent SOTA methods CSST-9stg and Restormer by 0.76 dB and 1.36 dB, demonstrating the effectiveness and acceptability of ADIS. Additionally, our models dramatically surpass SOTA methods while requiring lower computational and memory costs, and we obtain a significant gain on ADIS with only $29.96\%$ Params and $61.00\%$ FLOPS of Restormer and $69.05\%$ Params and $76.09\%$ FLOPS of CSST-9stg.

\subsubsection{Qualitative Comparisons}
\label{sec6.1.2}

Fig. \ref{Fig6} illustrates a comparative evaluation of simulated hyperspectral image (HSI) reconstructions using our ODAUVST-5stg method alongside other state-of-the-art (SOTA) approaches. An inset in the bottom-left corner provides magnified views of patches extracted from the full HSI, highlighting key differences.
Our ODAUVST-5stg method consistently delivers HSIs with exceptional visual fidelity, characterized by sharp textures and fine details that closely align with the ground truth (GT). In comparison, alternative methods fall short: reconstructions from ADIS acquisition often appear over-smoothed, obscuring subtle structural features, while exhibiting chromatic artifacts and speckle patterns absent in the GT.
To further validate our approach, we analyze intensity-wavelength spectral profiles at the heptagram-marked location in the RGB image (left). The profiles generated by ODAUVST-5stg show the strongest agreement with reference spectra, demonstrating superior spectral fidelity. These results emphasize the robustness of our framework for producing high-quality, spectrally consistent HSI reconstructions.

\subsection{Comparison with Other Spectral Imaging Systems}
\label{sec5.2}
To evaluate the effectiveness of ADIS with orthogonal diffraction encoding in SSI, we compare our method with two established approaches: a prism-based system~\cite{Aprism} and a DOE-based system~\cite{DOE_Jeon}.
As noted earlier, we exclude 10 HSIs of full resolution from the KAIST dataset during training of all reconstruction algorithms. Using these 10 HSIs as a test set, we simulate the image formation models of all three systems and conduct reconstruction for evaluating the systems' performance. Additionally, two full-resolution scenes are selected to visualize and compare various systems' reconstruction performance as shown in Fig.~\ref{Fig7}.
To ensure experimental fairness—balancing contributions from data engineering and deep learning, we reconstruct the prism-based system using the state-of-the-art RGB super-resolution algorithm MST++~\cite{MST++}, the DOE-based system with the authors' implementation~\cite{DOE_Jeon}, and ADIS via ODAUVST-5stg.
To standardize the physical setup, we set the sensor pixel size to 3.45 µm, matching our prototype system. Therefore, we resize the PSFs of the DOE-based and prism-based methods accordingly.

\vspace{-0.2cm}
\begin{table}[h]
\caption{Quantitative comparison of ADIS with DOE-based system and prism-based system, PSNR (dB) and SSIM are reported.}
\vspace{-0.5cm} 
\begin{center}
\resizebox{1\columnwidth}{!}{
\begin{tabular}{c|cccc}
\hline
\rowcolor[HTML]{EFEFEF} 
System                              & PSNR (dB)   & SSIM        & RMSE        & SAM         \\ \hline
\cellcolor[HTML]{EFEFEF}Prism-based~\cite{Aprism}   & 34.26          & 0.967          & 0.0199          & 4.45          \\
\cellcolor[HTML]{EFEFEF}DOE-based~\cite{DOE_Jeon} & 35.60          & 0.973          & 0.0173          & 4.34          \\
\cellcolor[HTML]{EFEFEF}ADIS (Ours) & \textbf{35.97} & \textbf{0.978} & \textbf{0.0169} & \textbf{3.96} \\ \hline
\end{tabular}
}
\end{center}
\label{tab2}
\vspace{-0.2cm} 
\end{table}

Table~\ref{tab2} presents the performance of three systems on the ten full-resolution scenes, evaluated using PSNR, SSIM, RMSE, and SAM as key metrics. Fig.~\ref{Fig7} illustrates that our imaging system achieves superior spatial and spectral reconstruction accuracy. Unlike the two compared methods, which rely on calibrated spectral encoding, our system employs a cost-effective binary mask whose diffraction effects can be accurately described physically without calibration, adding no footprint or operational complexity beyond that of standard RGB cameras.

\subsection{Comparison with Other Unfolding Frameworks}
To highlight ODAUF’s role in enhancing VST’s ability to exploit orthogonal diffraction degradation in ADIS, we compare ODAUVST’s reconstruction performance with two alternative frameworks: Jeon2019~\cite{DOE_Jeon}, which adapts deep unfolding from denoising to DOE-based spectral reconstruction, and COPF~\cite{ADIS}, an early realization of the ADIS reconstruction. By integrating VST as a denoiser across all frameworks with a fixed five-iteration limit, we assess PSNR, SSIM, memory costs and computational complexity, demonstrating ODAUF’s superior reconstruction efficiency.

\vspace{-0.2cm}
\begin{table}[h]
\caption{Quantitative comparison of ODAUF with two alternative frameworks for ADIS reconstruction.}
\vspace{-0.5cm} 
\begin{center}
\resizebox{1\columnwidth}{!}{
\begin{tabular}{c|cccc}
\hline
\rowcolor[HTML]{EFEFEF} 
System                              & PSNR (dB)   & SSIM        & Params        & GFLOPS         \\ \hline
\cellcolor[HTML]{EFEFEF}Jeon2019   & 30.33          & 0.907          & 3.79M          & 44.01          \\
\cellcolor[HTML]{EFEFEF}COPF & 31.92          & 0.937          & 3.85M          & 47.54          \\
\cellcolor[HTML]{EFEFEF}ODAUF & \textbf{33.31} & \textbf{0.951} & 4.53M & 53.64 \\ \hline
\end{tabular}
}
\end{center}
\label{tab3}
\vspace{-0.2cm} 
\end{table}

As shown in Table~\ref{tab3}, ODAUF significantly improves the performance of reconstruction without significantly increasing the overall memory and computational costs. Paired with VST, ODAUF enables a compact, calibration-free system design that surpasses the imaging quality of prior methods.

\begin{figure*}[t]
    \begin{center}
    \includegraphics[width=1\linewidth]{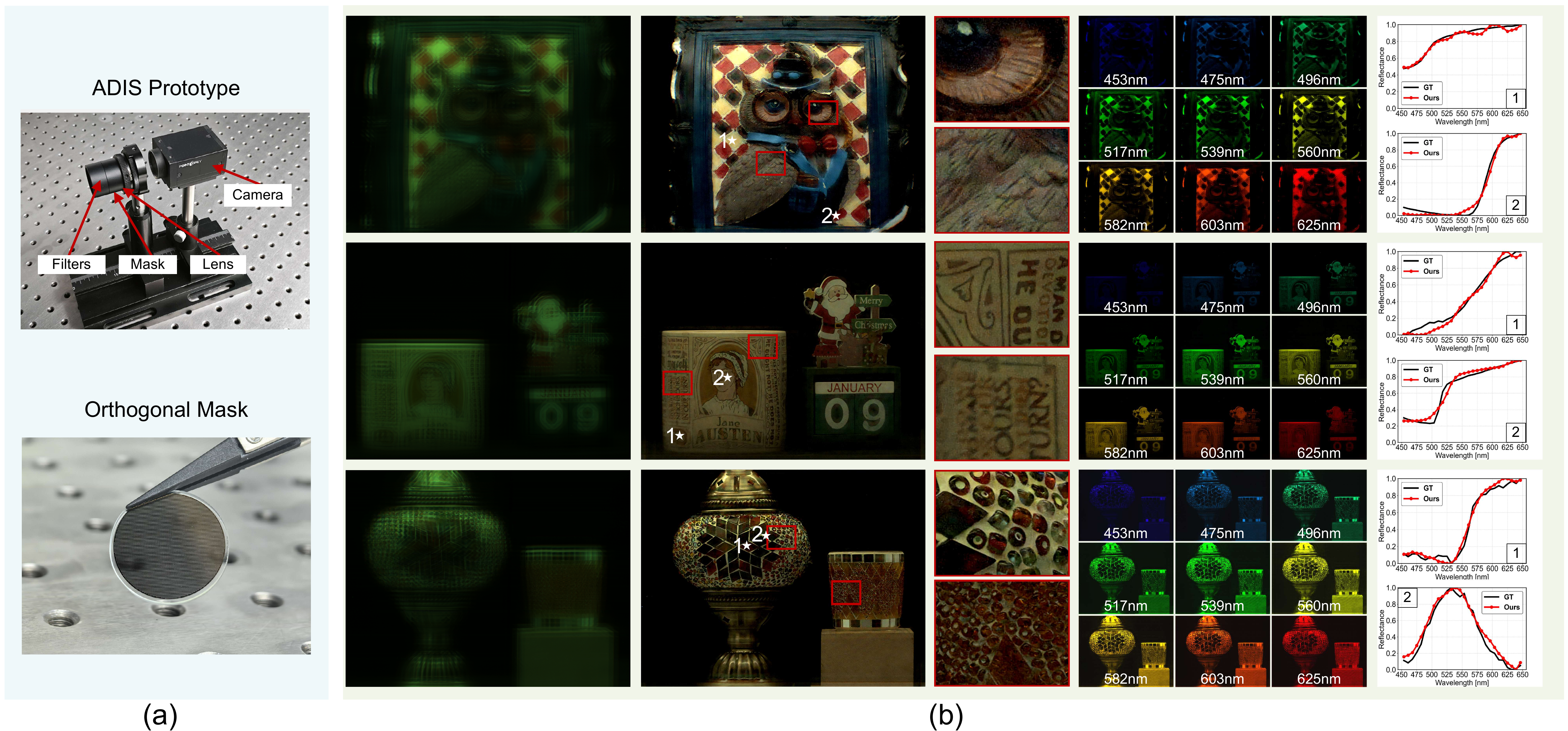}
    \end{center}
\vspace{-0.4cm} 
\caption{ADIS-V1: Compact prototype based on a single-lens configuration and representative reconstruction results. The mask is directly on the front surface of the lens. Unlike prior methods relying on calibrated PSFs, we reconstruct these scenes directly using computed PSFs. The reference spectral curves (ground truth, GT) are measured with a point spectrometer.}
\label{Fig8.1}
\vspace{-0.3cm} 
\end{figure*}

\subsection{System Noise Analysis}
\label{sec5.4}

Regarding system noise, here we quantitatively compare the performance of different systems and algorithms under varying noise levels. We not only demonstrate the robustness of ODAUVST to noise but also show that ADIS exhibits stronger resilience against noise interference compared to DOE-based methods.

To further assess the robustness of different reconstruction methods to measurement noise, we conducted additional experiments by adding zero-mean Gaussian noise with different standard deviations ($\varpi = 0.01$, $\varpi = 0.02$, and $\varpi = 0.05$) to the simulated measurements. Under these noise levels, we compared ODAUVST-5stg with Restormer and CSST-9stg, and the quantitative results are summarized in Table~\ref{newtab1.2.1}. 

\vspace{-0.2cm}
\begin{table}[h]
\caption{Quantitative comparison of ODAUVST-5stg with Restormer and CSST-9stg under different noise levels, PSNR (dB), and SSIM are reported.}
\vspace{-0.2cm} 
\begin{center}
\renewcommand{\arraystretch}{1.3}
\resizebox{1\columnwidth}{!}{
\begin{tabular}{c|ccc|ccc|ccc}
\hline
\rowcolor[HTML]{EFEFEF} 
\cellcolor[HTML]{EFEFEF}\textbf{Algorithm} & \cellcolor[HTML]{EFEFEF}\textbf{$\varpi$} & \cellcolor[HTML]{EFEFEF}\textbf{PSNR} & \textbf{SSIM}  & \textbf{$\varpi$} & \textbf{PSNR}  & \textbf{SSIM}  & \textbf{$\varpi$} & \textbf{PSNR}  & \textbf{SSIM}  \\ \hline
ODAUVST-5stg                               & 0.01                               & \textbf{32.38}                        & \textbf{0.928} & 0.02       & \textbf{31.62} & \textbf{0.908} & 0.05       & \textbf{30.11} & \textbf{0.873} \\
CSST-9stg~\cite{ADIS}                                  & 0.01                               & 32.12                                 & 0.923          & 0.02       & 31.34          & 0.906          & 0.05       & 30.06          & 0.872          \\
Restormer~\cite{restormer}                                  & 0.01                               & 31.30                                 & 0.916          & 0.02       & 30.51          & 0.897          & 0.05       & 29.44          & 0.869          \\ \hline
\end{tabular}
}
\end{center}
\label{newtab1.2.1}
\vspace{-0.3cm} 
\end{table}

We further compared the performance of ADIS and the DOE-based system under noise degradation. Specifically, we evaluated both methods with an additive noise standard deviation of $\varpi=0.01$, using the same experimental settings as in Section~\ref{sec5.2}. The results are summarized in Table~\ref{newtab1.2.2}.

\vspace{-0.2cm}
\begin{table}[h]
\caption{Quantitative comparison of ADIS with the DOE-based system under noise conditions.}
\vspace{-0.2cm} 
\begin{center}
\renewcommand{\arraystretch}{1.2}
\resizebox{1\columnwidth}{!}{
\begin{tabular}{c|cccl|cccc}
\hline
\rowcolor[HTML]{EFEFEF} 
\cellcolor[HTML]{EFEFEF}\textbf{System} & \cellcolor[HTML]{EFEFEF}\textbf{$\varpi$} & \cellcolor[HTML]{EFEFEF}\textbf{PSNR} & \textbf{SSIM}                         & \textbf{SAM}                         & \textbf{$\varpi$}    & \textbf{PSNR}                         & \textbf{SSIM}                         & \multicolumn{1}{l}{\cellcolor[HTML]{EFEFEF}\textbf{SAM}} \\ \hline
\textbf{ADIS}                           & 0.00                      & 35.97                        & 0.978                        & \multicolumn{1}{c|}{3.96}   & \textbf{0.01} & 32.81                        & 0.939                        & 5.90                                            \\
\textbf{DOE-based}                      & 0.00                      & {\color[HTML]{F56B00} \textbf{35.60}} & {\color[HTML]{F56B00} \textbf{0.973}} & {\color[HTML]{F56B00} \textbf{4.34}} & \textbf{0.01} & {\color[HTML]{F56B00} \textbf{30.35}} & {\color[HTML]{F56B00} \textbf{0.864}} & {\color[HTML]{F56B00} \textbf{6.36}}                     \\ \hline
\end{tabular}
}
\end{center}
\label{newtab1.2.2}
\vspace{-0.3cm} 
\end{table}

As shown in Table~\ref{newtab1.2.2}, the DOE-based approach exhibits a pronounced degradation in reconstruction quality even under such mild noise, whereas ADIS remains markedly more stable. This observation may also help explain why, in real data acquisition, DOE-based methods tend to produce noticeable white artifacts even after careful calibration.

\begin{figure*}[t]
    \begin{center}
    \includegraphics[width=0.98\linewidth]{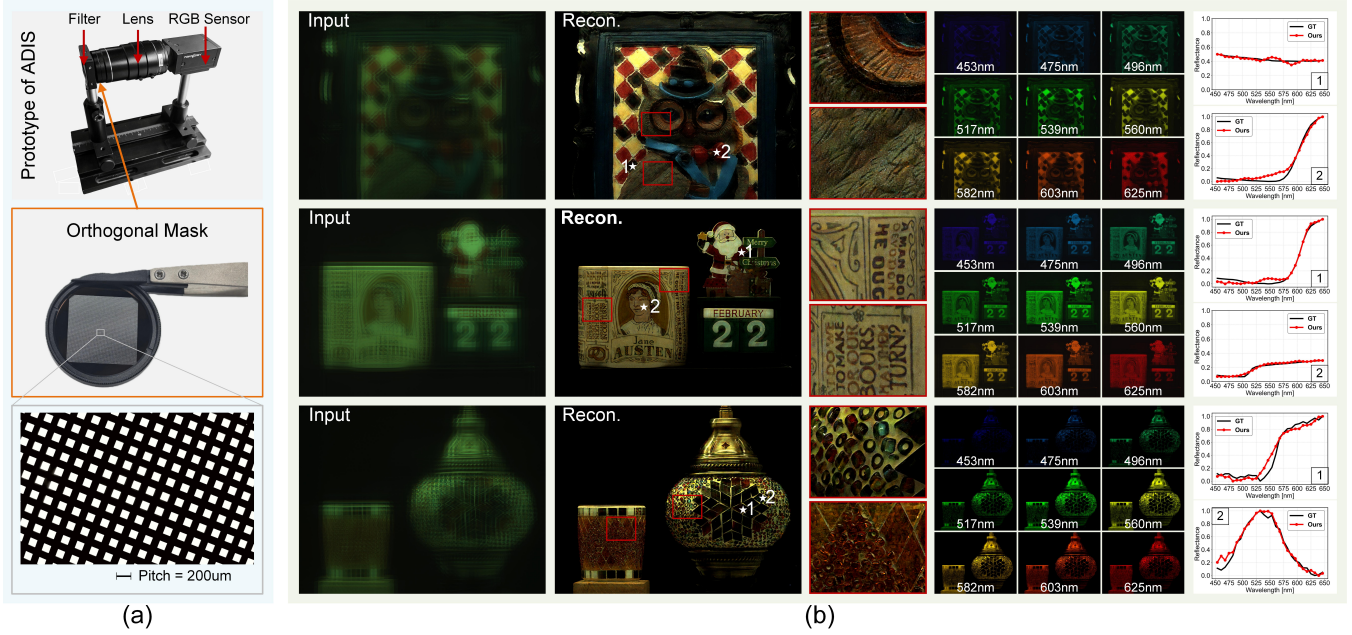}
    \end{center}
\vspace{-0.3cm} 
\caption{ADIS-V2: We capture three natural scenes using our real prototype system as shown in (a). We directly reconstruct these scenes utilizing the computed PSFs corresponding to those of ADIS-V1. The reference spectral curves (ground truth, GT) are measured with a point spectrometer. Notably, our method eliminates artifacts caused by simulation-to-reality gaps inherent to DOE-based acquisition~\cite{DOE_Jeon,wang2024non}, and the comparisons can be found in the Supp. Material.}
\label{Fig8.2}
\vspace{-0.2cm} 
\end{figure*}


\begin{figure*}[t]
    \begin{center}
    \includegraphics[width=1\linewidth]{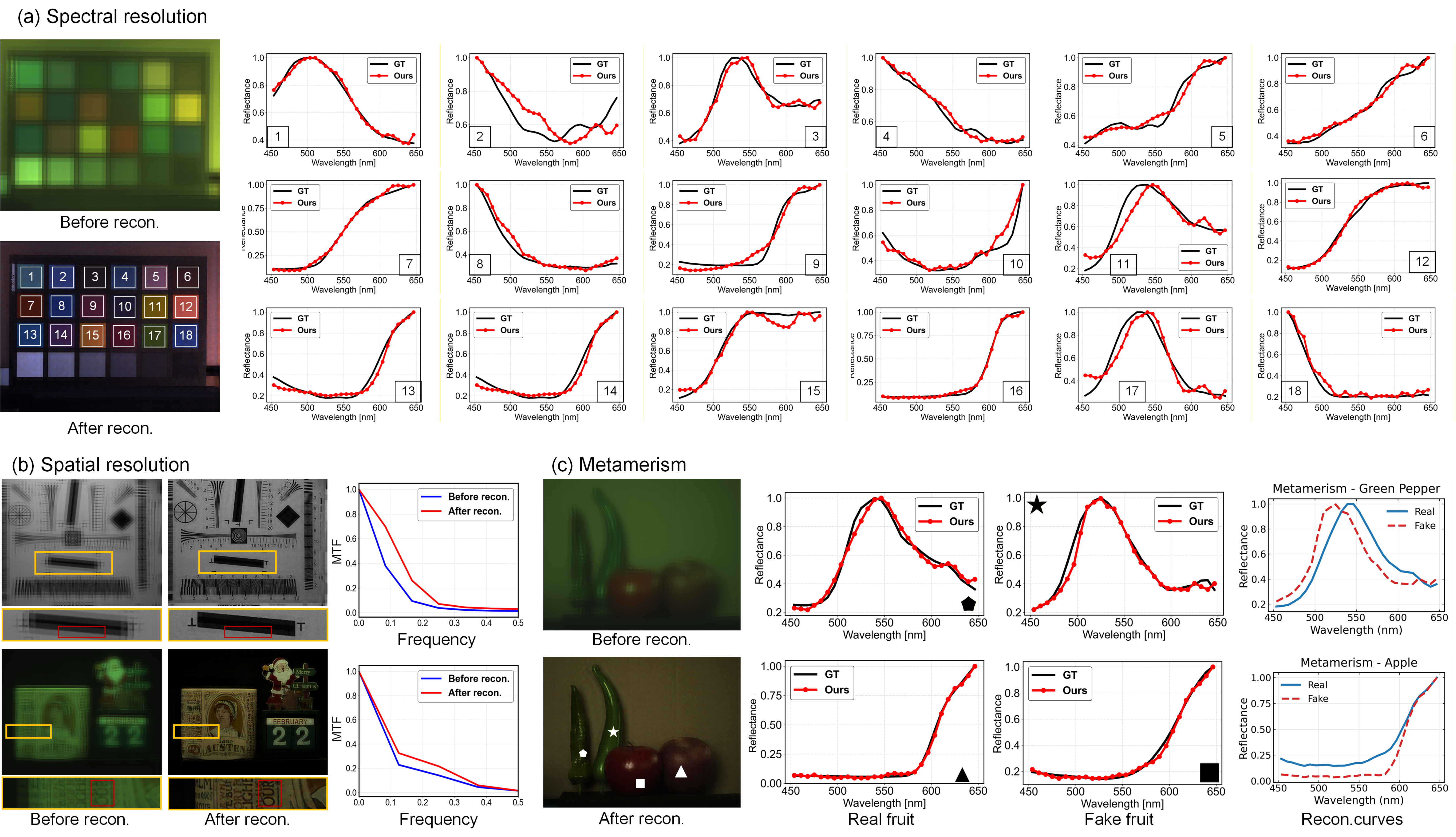}
    \end{center}
\vspace{-0.4cm} 
\caption{Quantitative evaluation of our ADIS prototype. (a) shows the reconstructed ColorChecker and reconstruction curves vs. ground truth. (b) demonstrates the spatial accuracy of our reconstruction. We compare the modulation transfer functions of the input image and the output reconstruction using the square region. (c) Metameric material discrimination using ADIS.}
\label{Fig9}
\vspace{-0.1cm} 
\end{figure*}

\begin{figure}[h]
    \begin{center}
    \includegraphics[width=1\linewidth]{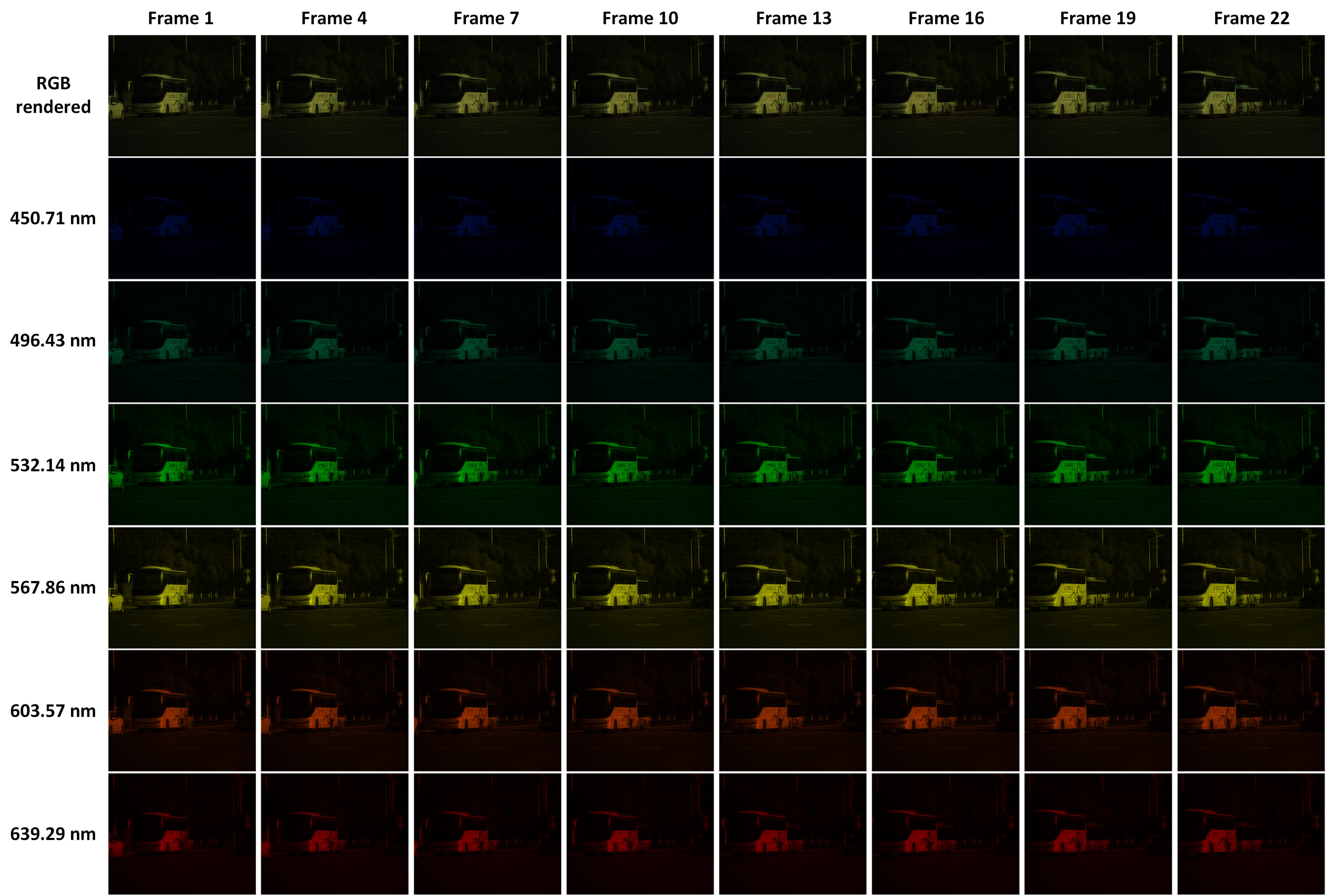}
    \end{center}
\vspace{-0.2cm} 
\caption{The snapshot spectral imaging results of ADIS-V2 in an autonomous-driving scenario. Please refer to the supplementary materials for the reconstruction video.}
\label{Fig9.2}
\vspace{-0.0cm} 
\end{figure}

\section{Real Experimental analysis}
\subsection{Implementation Details}
In our experiments, we build and evaluate two ADIS prototype systems. The first prototype, ADIS-V1 (Fig.~\ref{Fig8.1}, top left), employs a $50 mm$ focal-length singlet lens tightly bonded to an orthogonal binary mask with $25\%$ light throughput, forming the imaging diffractive lens (ODL), together with a standard RGB industrial camera (GS3-U3-51S5C). This configuration is primarily used to demonstrate the compactness of the ADIS architecture. The second prototype, ADIS-V2 (Fig.~\ref{Fig8.2}, top left), is constructed by tightly attaching a $25\%$-throughput mask to the front surface of a commercial $50 mm$ compound lens (Kowa LM50XC), again forming an ODL. For both prototype systems, we directly reuse the PSFs simulated for the $50 mm$ singlet to train a single reconstruction model, and then apply this model without any additional calibration, thereby validating the calibration-free property of ADIS across different lens assemblies. 

In both prototypes, an additional bandpass filter ($450–650 nm$) is inserted to restrict the operating spectral range, and an adjustable iris is included to control the aperture.
The mask, comprising two overlapping sets of $100\mu m$-wide parallel lines ($0.2 \mu m$ width uniformity), spans a $36.8 mm$-diameter photomask with a $23 \times 23$ $mm$ modulation area. 
It can be customized for $\$80$ per piece, with commercial mass production costing less than $\$1$ per unit. Once the physical setup of the system is determined, all projection relations can be easily calculated by Equation \ref{1-2}, even in the case of interference. Due to the simplicity of the construction and processing, all real experiments are carried out directly through the PSFs obtained from the theoretical calculations, and we only adjust the orientation of the orthogonal diffraction to be consistent with the theoretical PSFs by rotation. This straightforward design enables efficient capture of high-dimensional spectral information.

We train ODAUVST-5stg with the real configuration on CAVE-1024 and KAIST datasets jointly. Meanwhile, to address the disparity between real-world experiments and simulations arising from inherent noise and our omission of higher-order low-intensity diffraction, we incorporate randomized noise into the training data for model training, thereby bridging the aforementioned gaps.

\subsection{Qualitative and Quantitative Analysis}
\label{6.2}

The performance of real HSI reconstruction is demonstrated in Fig.~\ref{Fig8.1}(b) and Fig.~\ref{Fig8.2}(b), which present the measurements of a spatial size of $2448 \times 2048$ captured from real-world scenes, and the corresponding recovered spectral data of a spatial size of $2448 \times 2048 \times 28$. 
In terms of spatial structure, the reconstructed spectral data exhibit well-defined content, sharp textures, and minimal artifacts. Zoomed-in images of local details reveal that ADIS effectively captures spectral information while preserving spatial details and texture features at a high level.
Regarding spectral reconstruction accuracy, the predicted spectral curves at two spectrally significant points in each scene are compared with ground truth measurements obtained from a point spectrometer, demonstrating a high degree of agreement. The results presented here are grounded in a model trained with PSFs obtained from theoretical computations, providing substantial validation of the mathematical model, design framework, and reconstruction algorithm. In addition, they demonstrate that the system functions effectively without the need for calibration.

Notably, whether based on deep optics optimization~\cite{wang2024non} or forward design~\cite{DOE_Jeon} methodologies, DOE-based systems consistently suffer from artifacts in real-world imaging due to inherent simulation-to-reality discrepancies. Crucially, even systematic optimization or meticulous calibration fails to accurately model the optical modulation of physical DOEs. In contrast, ADIS employs a streamlined imaging pipeline that closely matches simulation, enabling calibration-free, high-fidelity snapshot spectral imaging with precision.

\subsection{Resolution Analysis}
\label{6.3}

\subsubsection{\textbf{Spectral accuracy}}
We captured a scene containing a standard ColorChecker illuminated by a D65 source to assess the spectral accuracy of ADIS reconstruction. Fig. \ref{Fig9}(a) demonstrates the performance of the real HSI reconstruction, presenting measured data from a real-world scene with a spatial resolution of $ 2448 \times 2048$ pixels alongside the corresponding reconstructed spectrum with a spatial resolution of $ 2448 \times 2048 \times 28$. The reconstructed spectral data exhibit well-defined structures and minimal artifacts.
Significantly, the predicted spectral curves for each color patch of the standard ColorChecker closely align with those obtained via a point spectrometer, confirming that the ADIS achieves a spectral resolution of less than $8 nm$ with the current configuration.

\subsubsection{\textbf{Spatial resolution}}
Fig. \ref{Fig9}(b) compares images of the standard resolution test chart (ISO12233) before and after spectral reconstruction, revealing a significant enhancement in the modulation transfer function (MTF) post-reconstruction. Similarly, we conducted an analogous comparison within a highly textured scene, displaying images across various spectral bands within the designated comparison area. This demonstrates our ability to distinctly resolve textures and structures in each band, effectively restoring the high-frequency details of the original data.

\subsection{Application Demonstration}
\label{6.3}

To further demonstrate the practical spectral imaging capability of ADIS, we conduct additional real-world experiments on metameric material discrimination and dynamic spectral video reconstruction.

\subsubsection{\textbf{Metameric Material Discrimination}}

Metameric materials exhibit similar RGB appearances but distinct spectral reflectance properties, providing a challenging test for spectral imaging systems. As shown in Fig.~\ref{Fig9}(c), ADIS successfully distinguishes different metameric materials by recovering their distinct spectral signatures. The reconstructed spectra closely agree with the reference measurements from a scanning spectral camera, demonstrating the capability of ADIS to preserve discriminative spectral information beyond RGB imaging.

\subsubsection{\textbf{Dynamic Spectral Video Reconstruction in Autonomous Driving Scenarios}}

We further evaluate ADIS in a dynamic outdoor environment using an ADIS-V2 prototype, which captures an autonomous-driving scene at 25 FPS. As shown in Fig.~\ref{Fig9.2}, ADIS continuously reconstructs spatial and spectral information with good spatio-temporal consistency throughout the video sequence. The corresponding spectral video results are provided in the supplementary material.

These experiments demonstrate the robustness of ADIS for real-world snapshot spectral imaging, covering both challenging metameric materials and dynamic outdoor scenarios.

\section{Ablation study and Hardware analysis}
In this section, we evaluate the LSM variants, the complete ODAUVST framework, and the impact of different mask designs on ADIS reconstruction performance. All experiments follow the setup in Section~\ref{sec6.1} and use ODAUVST-5stg for a fair comparison.

\begin{figure}[t]
    \begin{center}
    \includegraphics[width=1\linewidth]{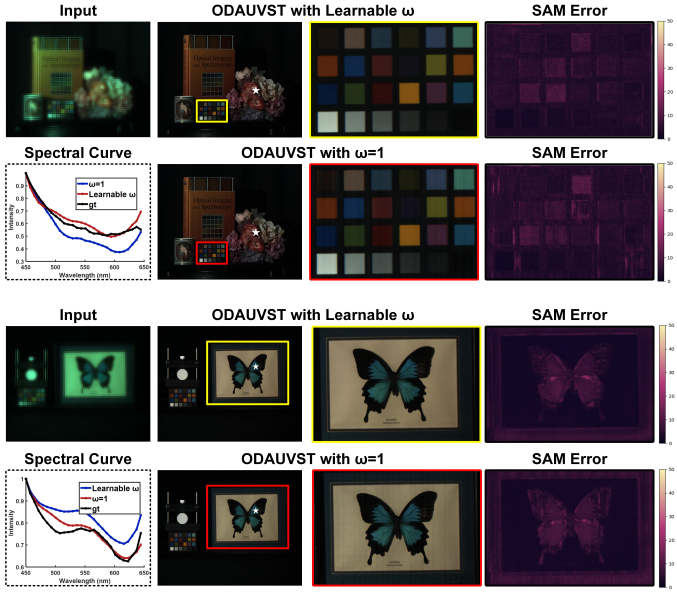}
    \end{center}
\caption{Performance comparison of ODAUVST-5stg with a fixed fractional shifts range $\omega$ and a learnable fractional shifts range $\omega$ on the test set.}
\label{Fig10}
\vspace{-0.1cm} 
\end{figure}

\vspace{-0.1cm}
\subsection{Ablation Study of LSM and CSM}
\label{7.1}

In the ODAUVST implementation, the fractional shifts computed by the learning shifts module (LSM) constitute a critical parameter, with the shift range directly influencing the performance of the vision system transformer (VST) in ADIS reconstruction.
In this study, the fractional shift range parameter $\omega$ within the LSM is designated as a learnable parameter. To substantiate this configuration, we evaluate the reconstruction performance across various $\omega$ values under consistent experimental conditions. Additionally, the fixed integer shifts $\mathcal{S}$ also influence the feature extraction capability of the VST in orthogonal diffraction perception modes. We further assess the performance of ODAUVST-5stg with fixed shift steps of [1, 1], [2, 2], and [3, 3] when 
$\omega$ is learnable, revealing that the fixed step size of [1, 1] yields optimal performance—both with $\omega = 1$ and as a learnable parameter.

\begin{table}[h]
\caption{Ablation study on LSM with different fixed integer shifts $\mathcal{S}$ and different fractional shifts range $\omega$ .}
\vspace{-0.3cm} 
\begin{center}
\renewcommand{\arraystretch}{1.2}
\resizebox{1\columnwidth}{!}{
\begin{tabular}{c|ccc|ccc}
\hline
\rowcolor[HTML]{EFEFEF} 
Fixed integer shifts $\mathcal{S}$             & {[}1,1{]} & {[}1,1{]} & {[}1,1{]}                             & {[}1,1{]}                             & {[}2,2{]} & {[}3,3{]} \\
\rowcolor[HTML]{EFEFEF} 
Fractional shifts range $\omega$         & 3         & 2         & 1                                     & learned                               & learned   & learned   \\ \hline
\rowcolor[HTML]{FFFFFF} 
\cellcolor[HTML]{EFEFEF}PSNR/dB & 33.01     & 33.14     & {\color[HTML]{CB0000} \textbf{33.35}} & {\color[HTML]{3531FF} \textbf{33.31}} & 31.83     & 30.99     \\
\rowcolor[HTML]{FFFFFF} 
\cellcolor[HTML]{EFEFEF}SSIM    & 0.944     & 0.947     & {\color[HTML]{3531FF} \textbf{0.949}} & {\color[HTML]{CB0000} \textbf{0.951}} & 0.929     & 0.912     \\ \hline
\end{tabular}
}
\end{center}
\label{tab4}
\vspace{-0.2cm} 
\end{table}

In our experiments, a notable phenomenon emerges, as evidenced in Table~\ref{tab4}: setting the fractional shifts range $\omega$ as 1 yields comparable reconstruction performance compared to setting  $\omega$ as a learnable parameter. To refine the configuration strategy for 
$\omega$, we evaluate the generalizability of these two implementations on full-resolution scenes, as depicted in Fig. \ref{Fig10}. Although a fixed range can produce results akin to a learnable $\omega$ by overfitting within a limited test range, it significantly compromises model generalizability, hindering real-scene reconstruction. Consequently, we adopt the learnable $\omega$ as the optimal configuration for the LSM. In the later section \ref{sec8}, we further discuss how this behaves in real experiments.

\vspace{-0.2cm}
\begin{table}[h]
\caption{Ablation study on CSM with different padding methods.}
\vspace{-0.2cm} 
\begin{center}
\resizebox{0.9\columnwidth}{!}{
\begin{tabular}{c|ccc}
\hline
\rowcolor[HTML]{EFEFEF} 
\cellcolor[HTML]{EFEFEF}\textbf{padding modes}               & reflection                   & zero padding                 & border                       \\ \hline
\cellcolor[HTML]{EFEFEF}{\color[HTML]{000000} PSNR} & {\color[HTML]{F56B00} 33.31} & {\color[HTML]{000000} 33.10} & {\color[HTML]{000000} 32.20} \\
\cellcolor[HTML]{EFEFEF}{\color[HTML]{000000} SSIM} & {\color[HTML]{F56B00} 0.951} & {\color[HTML]{000000} 0.948} & {\color[HTML]{000000} 0.945} \\ \hline
\end{tabular}
}
\end{center}
\label{tabadd}
\vspace{-0.2cm} 
\end{table}

In the CSM, we adopt reflection padding. To justify this design choice, we compare three padding modes-reflection, zero padding, and border. As shown in Table~\ref{tabadd}, reflection padding delivers the best reconstruction performance, as it effectively avoids boundary artifacts and preserves edge continuity, leading to overall superior imaging quality.

\subsection{Ablation Study of ODAUVST}
\label{7.2}

ODAUVST assists ADIS in achieving optimal compact SSI performance by combining ODAUF and VST. In order to demonstrate the critical role of ODAUF and VST as important components of the algorithm, we align and compare them with the key components of the CSST, which consists of COPF and SST in the previous version, facilitating a comprehensive ablation study (noting that VST is derived by incorporating learnable fractional shifts into SST).

\begin{table}[h]
\caption{Ablation study on ODAUF, VST with COPF, SST}
\vspace{-0.5cm} 
\begin{center}
\renewcommand{\arraystretch}{1.2}
\resizebox{1\columnwidth}{!}{
\begin{tabular}{ccc|ccccc}
\hline
\rowcolor[HTML]{EFEFEF} 
Framework & Denoiser & K & PSNR/dB        & SSIM           & Time & Params & GFLOPs \\ \hline
\rowcolor[HTML]{FFFFFF} 
COPF      & SST      & 5 & 31.58             & 0.936             & 5ms   & 3.67M     & 40.49     \\
\rowcolor[HTML]{FFFFFF} 
ODAUF     & SST      & 5 & 32.97             & 0.946             & 12ms   & 4.34M     & 46.59     \\
\rowcolor[HTML]{FFFFFF} 
COPF      & VST      & 5 & 31.92          & 0.937          & 16ms & 3.85M  & 47.54  \\
\rowcolor[HTML]{FFFFFF} 
ODAUF     & VST      & 5 & \textbf{33.31} & \textbf{0.951} & 17ms & 4.53M  & 53.64  \\
\rowcolor[HTML]{FFFFFF} 
COPF      & SST      & 9 & 32.55          & 0.945          & 17ms & 6.56M  & 70.50  \\ \hline
\end{tabular}
}
\end{center}
\label{tab5}
\vspace{-0.2cm} 
\end{table}

As presented in Table~\ref{tab5}, with the number of iterations set to $K=5$, the integration of the ODAUF and VST yields reconstruction performance improvements over the COPF and SST combination (i.e., CSST), achieving gains of 1.73dB in PSNR and 0.015 in SSIM. Additionally, when comparing CSST-9stg to ODAUVST-5stg, the latter demonstrates superior performance—0.76dB in PSNR and 0.015 in SSIM—while utilizing only $69.05\%$ of the parameters and $76.09\%$ FLOPS of CSST-9stg. 
Compared to CSST, standalone implementations of ODAUF and VST result in significant improvements of 1.39dB in PSNR, 0.014 in SSIM and 0.34dB in PSNR, 0.005 in SSIM, respectively. Thus, ODAUVST outperforms CSST in sensing orthogonal diffractive degradation patterns and achieves more efficient ADIS reconstruction.

\subsection{Comparison between different forms of masks}
\label{7.3}

In this study, we develop the prototype for the ADIS by extending the fundamental diffraction framework of the computational tomography imaging spectrometer (CTIS) and adopting a uniform diffraction blending approach, resulting in a system that requires no calibration due to its simple, homogeneous mask design. The ensuing question is whether the mask form can affect the reconstruction results of ADIS. For this reason, we simulate the diffraction forms of several different masks: quadrilateral, triangular, pentagram, and circular, and carry out simulation reconstruction experiments with a consistent image generation model.

\begin{figure}[t]
    \begin{center}
    \includegraphics[width=1\linewidth]{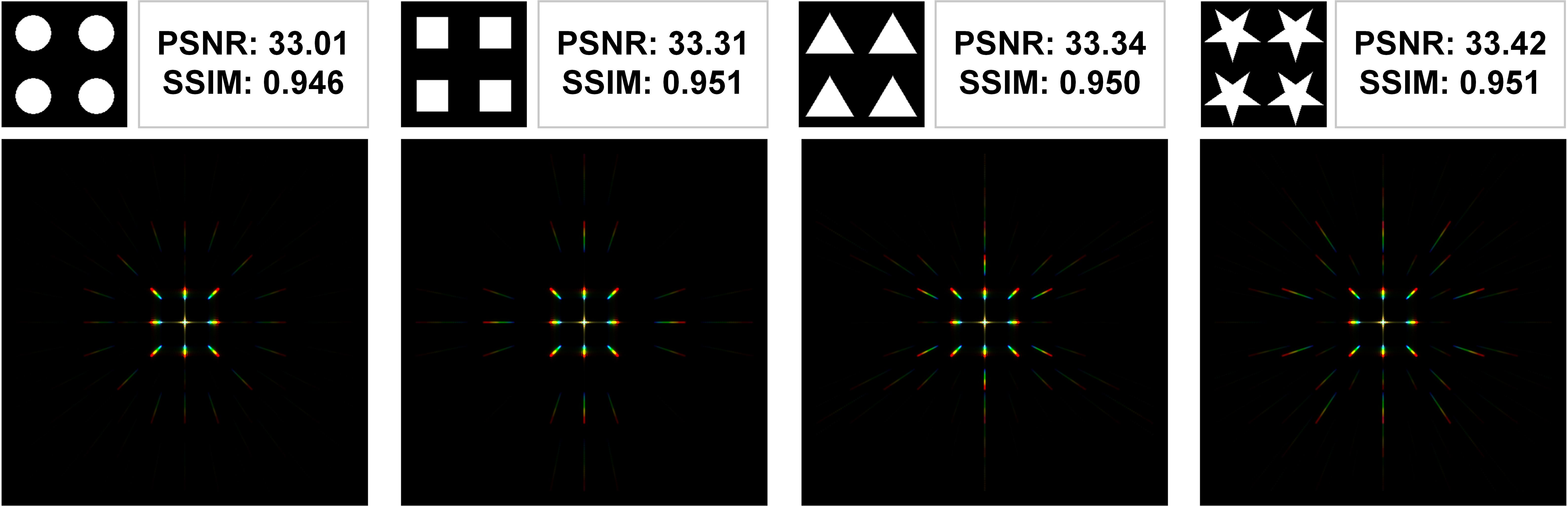}
    \end{center}
\vspace{-0.2cm} 
\caption{Comparison of simulation reconstruction with different mask forms. }
\label{Fig11}
\vspace{-0.1cm} 
\end{figure}

Fig. \ref{Fig11} presents the experimental results, revealing that the asymmetric PSF enables triangular and pentagram masks to achieve marginally higher reconstruction accuracy in simulations. However, this asymmetry complicates rotational adjustment of diffraction direction during application and increases the complexity and cost of mask fabrication via laser direct writing, particularly at acute angles, thereby widening the disparity between simulation and real-world outcomes. Consequently, these modest accuracy gains may not translate to practical experiments. In contrast, the centrally symmetric circular hole mask proves slightly less effective than the square-hole mask in empirical tests. Thus, the square-hole array mask remains a practical and effective choice for ADIS.

\section{Discussion}
\label{sec8}

\subsection{Spatial Variation of PSF}
\label{sec8.1}
\begin{figure}[t]
    \begin{center}
    \includegraphics[width=1\linewidth]{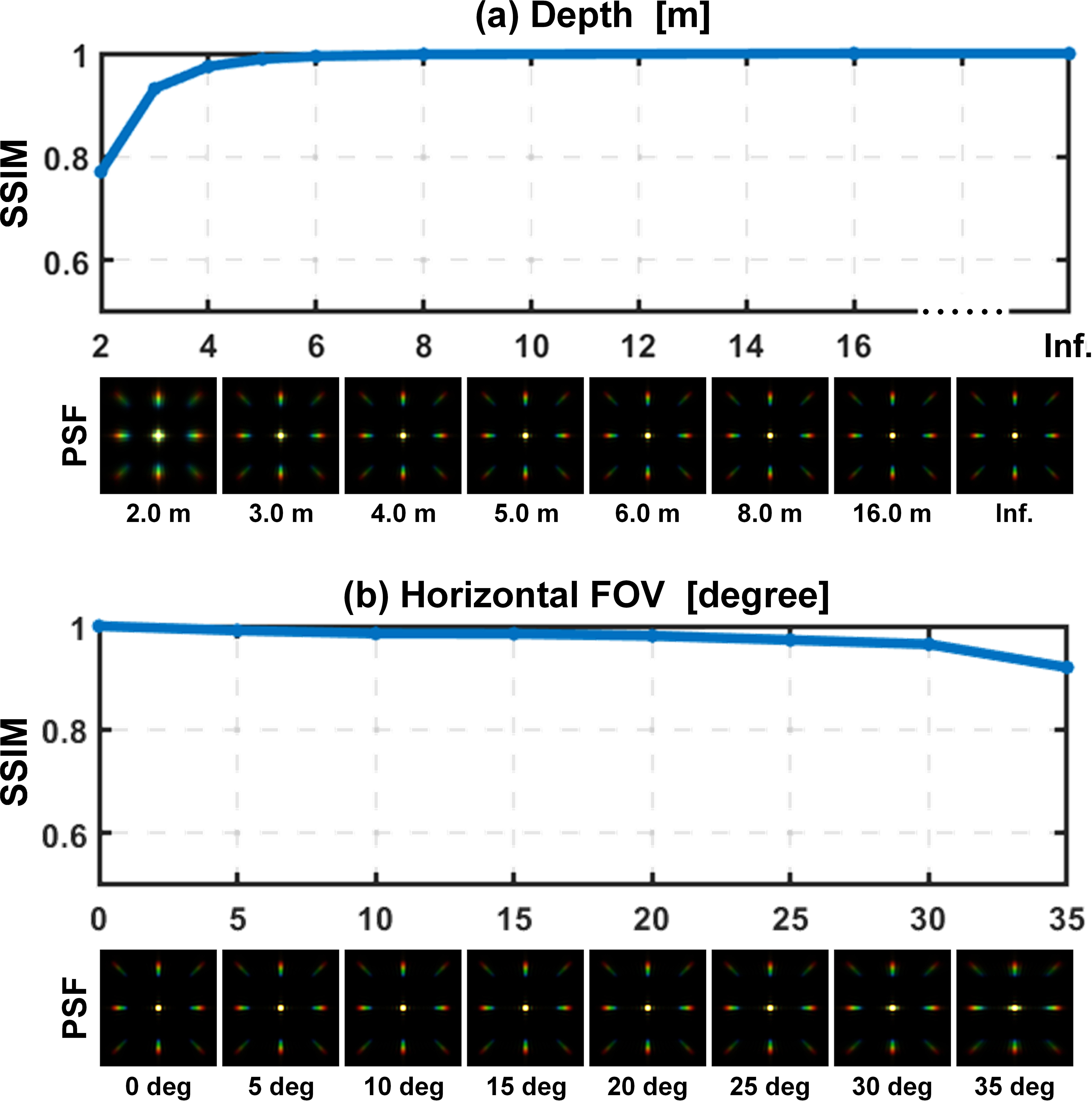}
    \end{center}
\vspace{-0.2cm} 
\caption{(a) The depth invariance of ADIS is observed when the depth is longer than 3 m; i.e., the SSIM of PSFs increases significantly (higher than 0.9). (b) The PSFs at different positions from the reference center to the end of the vertical FOV are compared within the valid field of view of 35 degrees. The position variance of the PSF is negligible with our optical configuration.}
\label{Fig12}
\vspace{-0.3cm} 
\end{figure}

\textbf{Depth dependency.}
In Section~\ref{sec3.2}, we demonstrated that under far-field imaging conditions, ADIS exhibits depth-invariant PSFs. To further validate this invariance, we configure ADIS to focus at infinity and simulate point sources at varying depths, ranging from $2.0 m$ to optical infinity, to examine PSF variations. As shown in Fig. \ref{Fig12}(a), we present the PSF shapes and SSIM indices at different depths, using the PSF at optical infinity as the reference. The results confirm that PSF depth variance is negligible beyond $3.0 m$ when the imaging system is focused at infinity, indicating that the PSF generated by the ODL primarily depends on wavelength. Additionally, for close-range imaging, adjusting the lens to focus at a specific depth enables effective SSI.

\vspace{0.1cm}
\textbf{Position dependency.}
The morphology of the point spread function (PSF) varies with the position of the point source within the scene, beyond its primary dependence on factors such as mask geometry, lens focal length, and object depth. Specifically, the $x,y$ coordinates of the source alter the PSF, constituting a key factor that constrains the imaging performance of most systems in wide-angle configurations. In both simulation and empirical experiments, we assume spatial invariance of the PSF in the augmented diffraction imaging system (ADIS) across $x,y$ positional changes. To examine the impact of spatial position on PSF morphology, we analyze the PSF of ADIS at varying distances from the reference center to a 35-degree viewing angle. As illustrated in Fig. \ref{Fig12}(b), the PSF diminishes gradually as the spatial position deviates from the optical center, yet the SSIM maintains a consistent performance above 0.9, even at the 35-degree edge. Consequently, the influence of position on PSF shape appears negligible, suggesting that ADIS, with appropriate lens optimization, holds significant potential for wide-angle spectral imaging.

\subsection{Lower Simulation-to-Reality Gap}
\label{sec8.2}

The discrepancy between simulation and real-world acquisition remains a major challenge for snapshot spectral imaging. Existing approaches, such as CASSI and DOE-based systems, typically rely on precise calibration of optical parameters and encoding patterns. However, practical factors including mechanical perturbations and environmental variations can introduce model mismatch, degrading reconstruction performance.

In contrast, ADIS employs a simple and easily characterized binary mask for spectral encoding and leverages theoretically computed PSFs for calibration-free reconstruction. By tolerating moderate lens-dependent optical variations, ADIS achieves a more seamless transition from simulation to real-world imaging, enabling reliable spectral reconstruction with reduced performance degradation.

\begin{figure}[t]
    \begin{center}
    \includegraphics[width=1\linewidth]{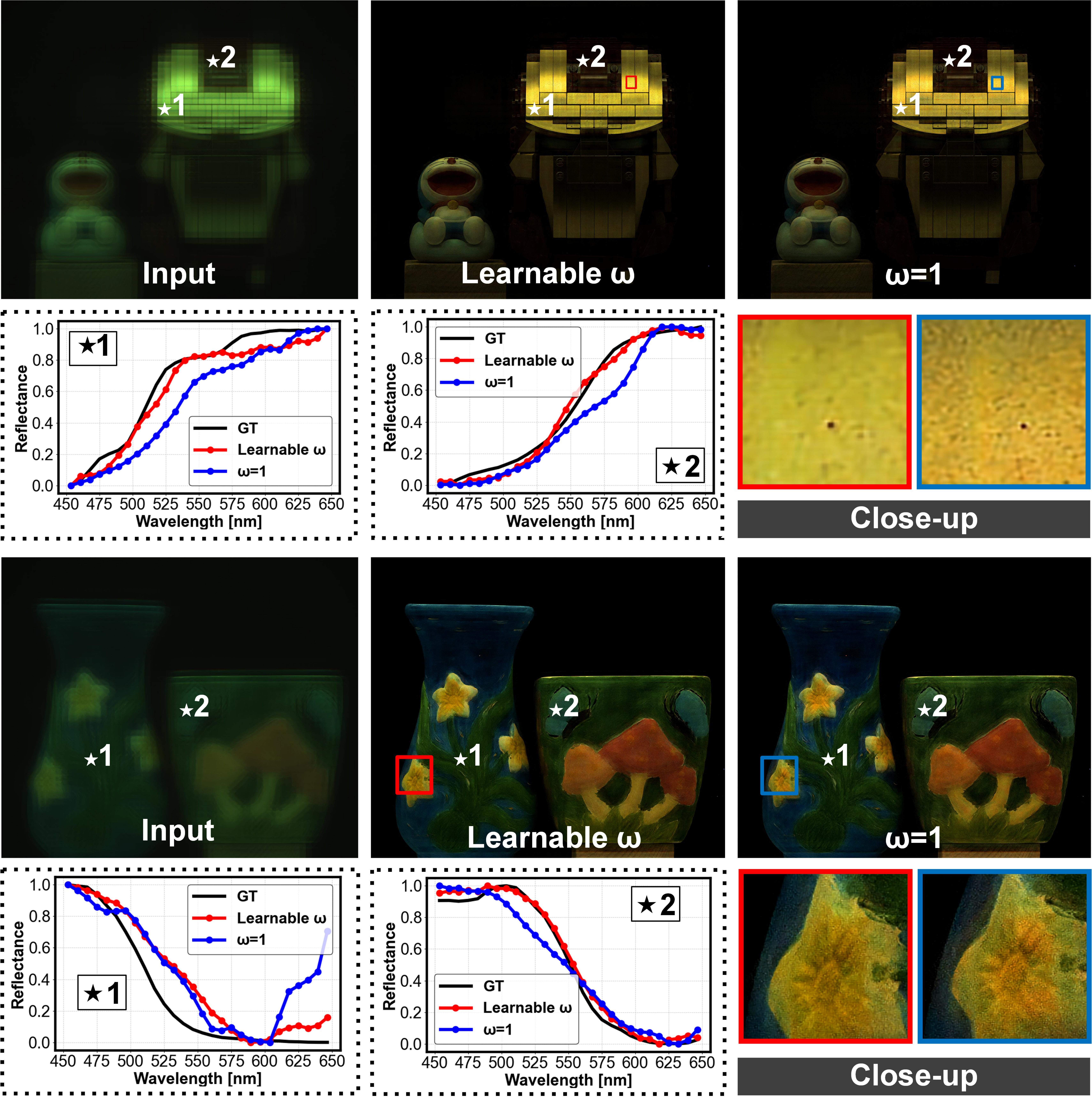}
    \end{center}
\caption{ADIS with ODAUVST (learnable $\omega$) achieves high-quality real spectral acquisition in scenes with smooth content.}
\label{Fig13}
\end{figure}

\subsection{Lower Edge Dependency}
From a diffraction coding perspective, high-dimensional spectral reconstruction relies on the inter-spectral discretization of PSFs. However, DOE-based approaches distribute PSFs over a larger area for diffraction coding, significantly degrading high-frequency image structures. As a result, distinguishable spectral information is primarily concentrated at the edges, leading to a strong edge dependence.
In contrast, ADIS exhibits relatively strong zero-order diffraction, which effectively preserves the original scene structure. Meanwhile, textures and edge details undergo multi-directional dispersion, enabling a richer and more explicit encoding. These properties reduce the system’s dependence on edge richness for spectral information and image structure recovery.

As shown in Fig. \ref{Fig8.1}, Fig. \ref{Fig8.2} and Fig. \ref{Fig13}, our method consistently achieves high-quality reconstructions in both edge-rich regions and smooth-content areas.
ODAUVST, with a learnable range of fractional shifts, enhances the generalization capability of ADIS reconstruction in real-world scenarios. This reduces edge dependence from both an algorithmic and coding perspective, allowing our method to be more adaptable across diverse scenes.

\subsection{Discussion on Calibration-Free Characteristics }
\label{sec8.4}
Here, we discuss the calibration-free nature of ADIS from two complementary perspectives: fabrication and optical aberrations:

\subsubsection{Calibration-Free Characteristics Regarding Lens Aberrations}

To validate this, we compare measured PSFs obtained under different lens configurations (compound lens, cemented achromatic lens and biconvex lens) with the simulated PSFs under the same conditions ($f=50 mm$, $d=200 \mu m$). As can be seen in Fig. \ref{Fig16}(a), as long as the single or compound lens shares the same effective focal length - thus enforcing the same diffraction propagation distance - the relative positions of the PSF lattice points remain nearly identical across different lens designs. This demonstrates that ADIS is intrinsically robust to the differing optical aberrations introduced by various lens configurations. On the fabrication side, ADIS further benefits from the fact that laser lithography can realize the designed mask pattern with high fidelity (fabrication error $< 0.2 \mu m$), so that the measured PSF lattices under all lens configurations closely match the simulated ones, reinforcing this robustness.

\subsubsection{Calibration-Free Characteristics Regarding Fabrication Errors of Mask}

\begin{figure*}[t]
    \begin{center}

    \includegraphics[width=1\linewidth]{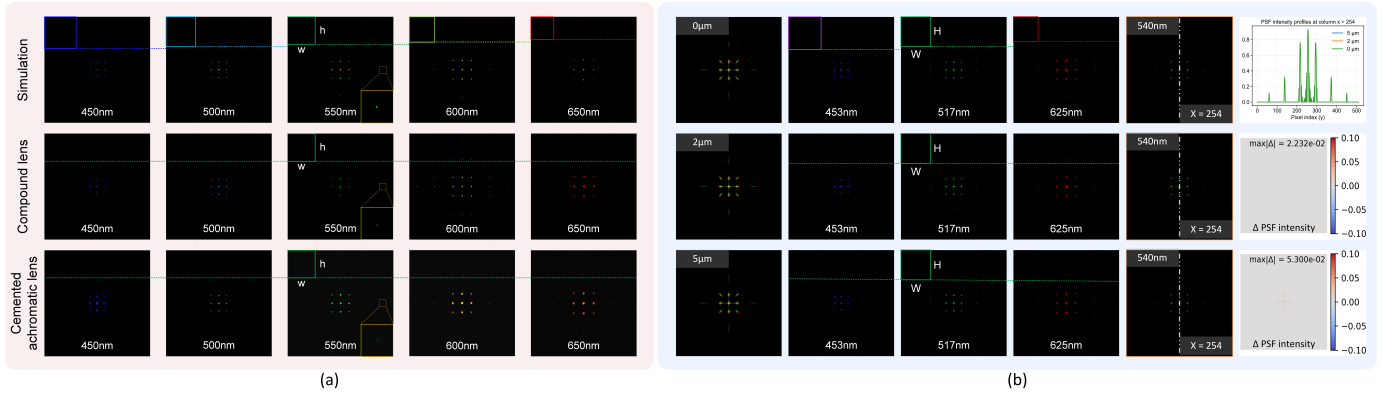}
    \end{center}
\vspace{-0.3cm} 
\caption{(a) Comparison of simulated PSFs generated by an ADIS mask with 100 $\mu m$ line width and 200 $\mu m$ period under different random fabrication errors (0, 2, 5 $\mu m$). (b) Comparison between theoretically calculated PSFs and experimentally measured ADIS PSFs under different lens configurations sharing the same effective focal length. For additional PSFs of ADIS under different random errors and different lenses, please refer to the supplementary materials.}
\label{Fig16}
\vspace{-0.1cm} 
\end{figure*}

Through further examining the impact of fabrication errors on the ADIS mask design, we find that the masks used in ADIS are not only straightforward to fabricate, but also inherently robust to random manufacturing deviations from a diffraction standpoint.
Specifically, for a representative design with a line width of  $100 \mu m$ and a period of  $200 \mu m$, we introduce random fabrication errors with amplitudes of $2 \mu m$ and $5 \mu m$. Even under these perturbations, the simulated PSFs produced by ADIS preserve a high degree of structural consistency, as shown in Figure \ref{Fig16}(b). This indicates that the diffractive masks employed in ADIS are both fabrication-friendly and resilient to realistic manufacturing tolerances.

From a fabrication perspective, ADIS enjoys inherent advantages: standard laser lithography can reproduce the designed mask with high fidelity (fabrication errors $< 0.2 \mu m$), and the ADIS design itself exhibits substantial tolerance to residual manufacturing deviations. Together, these properties give ADIS a naturally calibration-free character at the hardware level.

\subsection{Limitations}
\label{sec8.5}

\subsubsection{\textbf{Light throughput}}
ADIS employs minimal amplitude diffraction coding to enable calibration-free snapshot spectral imaging, delivering high-quality results in real-world applications. However, the attenuation of incident light due to amplitude coding reduces the overall light throughput of the imaging system. A potential solution, based on the Babinet principle (Fig. \ref{fig2}(a)), involves using complementary mask designs to generate similar PSFs for spectral reconstruction while retaining $75\%$ of the light throughput. Although this approach maintains comparable imaging performance, the reduction in diffraction efficiency may increase reliance on filtering, potentially compromising spectral fidelity in unknown scenes. Therefore, how to realize enhanced light throughput without degrading overall performance is one of the issues worth investigating.

\subsubsection{\textbf{Illumination}}
In our experiments, we found that the reconstructed spectra exhibit possible distortion when the light source spectrum of the illuminated scene is very complex or possesses high-frequency characteristics, because the training dataset uses CAVE and KAIST, both of which are reflectance data. While this distortion can be mitigated by dot-multiplying the light source spectrum with the dataset to enhance applicability in specific scenarios, its effectiveness remains limited under complex mixed lighting conditions.

\section{Conclusion}
We propose a diffraction-based compact snapshot spectral imaging (SSI) method using a simple orthogonal mask, the Aperture Diffraction Imaging Spectrometer (ADIS). ADIS enables direct migration of a model trained on theoretically computed PSFs to real-world acquisition, tolerating lens-dependent variations across different optical configurations, achieving compact SSI without calibration and significantly bridging the gap between simulation and real acquisition. ADIS maintains a consistent diffraction encoding with spatial invariance across the object space, making it well-suited for high-resolution, wide field-of-view applications.
Building on the system design, we develop ODAUVST, which enhances the perception of orthogonal diffraction patterns through learnable fractional shifts. By incorporating a learnable fractional shift range, ODAUVST improves generalization, ensuring that ADIS remains calibration-free while delivering consistently high imaging performance across both smooth and complex textures and structures. 
Synthetic simulation and physical experiments have verified the effectiveness of
our method in compact calibration-free SSI with minimal hardware cost.


\section*{Acknowledgments}
This research was supported by the National Key Research and Development Program of China (2023YFF0713300), the Fundamental Research Funds for the Central Universities (020414380175), and the Major Program of Science and Technology Project of Jiangsu Province (SBG2024000145).

\bibliographystyle{unsrt}
\bibliography{ADIS_PAMI}


\newpage

\section{Biography Section}

\begin{IEEEbiography}[{\includegraphics[width=1in,height=1.25in,clip,keepaspectratio]{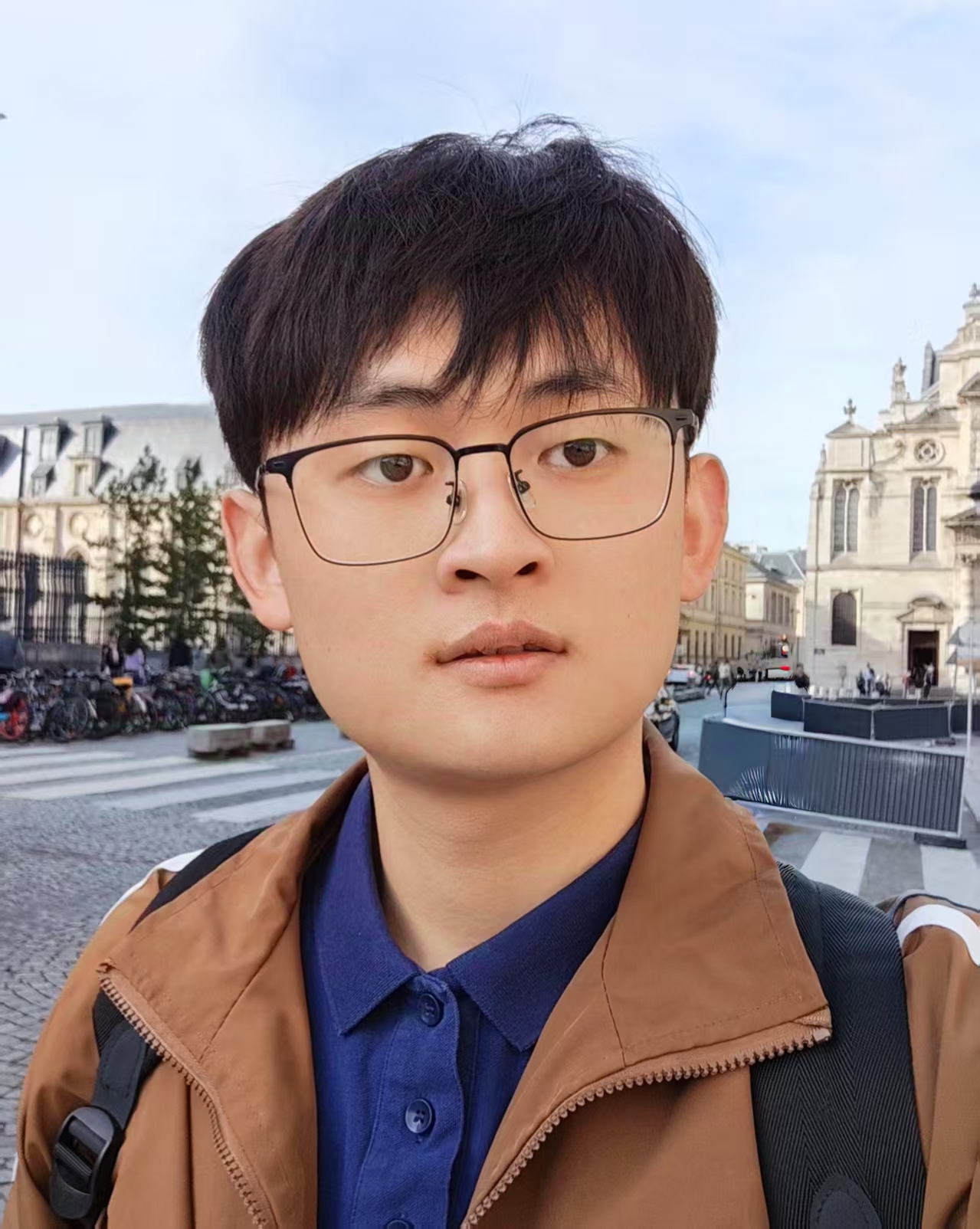}}]{Tao Lv} received the BS degree from the College of Information Science and Engineering, Northeastern University, Liaoning, China, in 2021. He is currently working toward the PhD degree with the School of Electronic Science and Engineering, Nanjing University, Nanjing, China. His research interests include computational photography and computer vision, especially computational spectral imaging.
\end{IEEEbiography}
\vspace{11pt}

\begin{IEEEbiography}[{\includegraphics[width=1in,height=1.25in,clip,keepaspectratio]{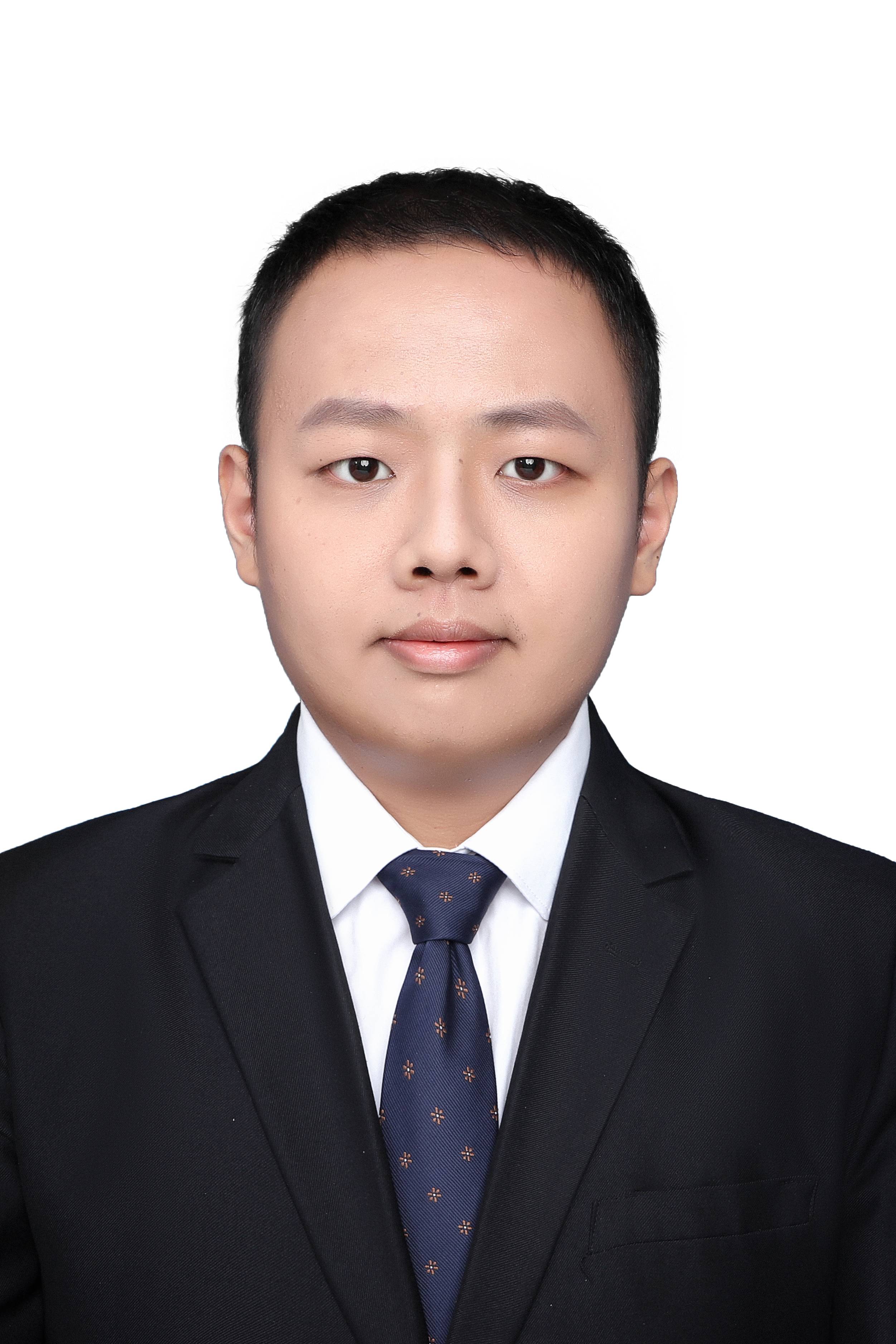}}]{Quan Yuan} received the B.S. degree in materials physics from Northwestern Polytechnical University, Xi’an, China, in 2018, and the PhD degree in optics from the School of Physics, Nanjing University, Nanjing, China, in 2024. Since then, he is currently an associate research fellow in the School of Physics at Nanjing University. His research interests include metasurface optics, micro- and nano-optics, and spectral imaging.
\end{IEEEbiography}
\vspace{11pt}

\begin{IEEEbiography}[{\includegraphics[width=1in,height=1.25in,clip,keepaspectratio]{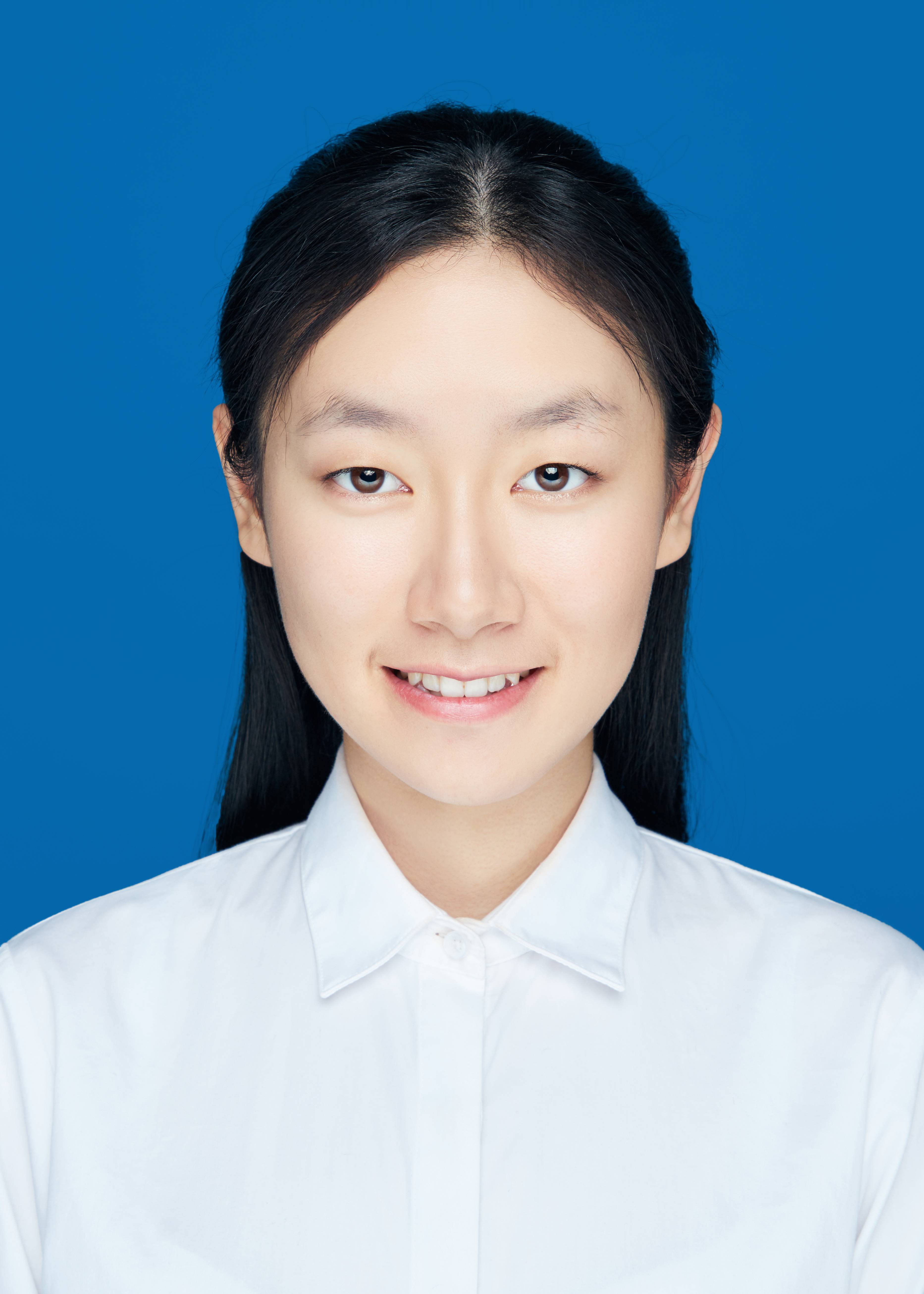}}]{Shiqiao Li} received the BS degree in Electronic Information Engineering from Nanjing University, Nanjing, China, in 2023. She is currently pursuing the M.Sc. degree in the School of Electronic Science and Engineering at Nanjing University, under the supervision of Prof. Xun Cao. Her research interests include computational imaging, spectral light field reconstruction, and high-dimensional image super-resolution.
\end{IEEEbiography}
\vspace{11pt}

\begin{IEEEbiography}[{\includegraphics[width=1in,height=1.25in,clip,keepaspectratio]{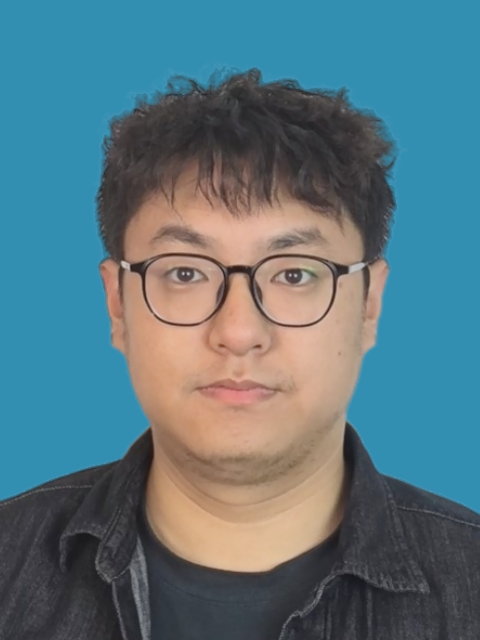}}]{Chenglong Huang} received the BS degree in communication engineering from Hohai University, China, in 2023. He is a graduate student from the School of Electronic Science and Engineering, Nanjing University.
\end{IEEEbiography}
\vspace{11pt}

\begin{IEEEbiography}[{\includegraphics[width=1in,height=1.25in,clip,keepaspectratio]{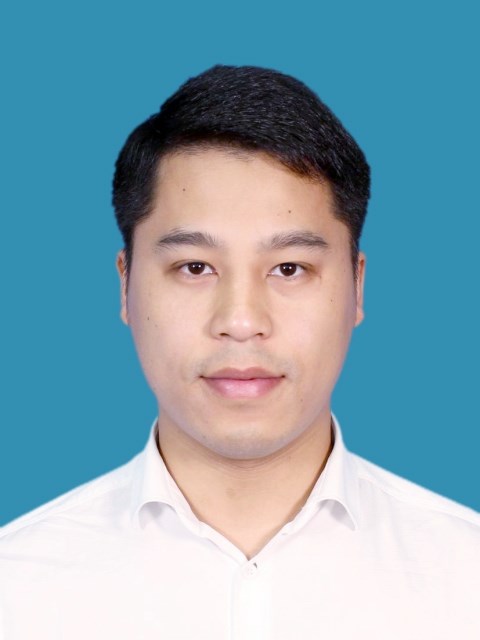}}]{Linsen Chen}
received the BS, MS and PhD degrees in 2014, 2017 and 2023 from School of Electronic Science and Engineering, Nanjing University, Nanjing, China. His research interests include computational photography, spectral imaging and reconstruction.
\end{IEEEbiography}
\vspace{11pt}

\begin{IEEEbiography}[{\includegraphics[width=1in,height=1.25in,clip,keepaspectratio]{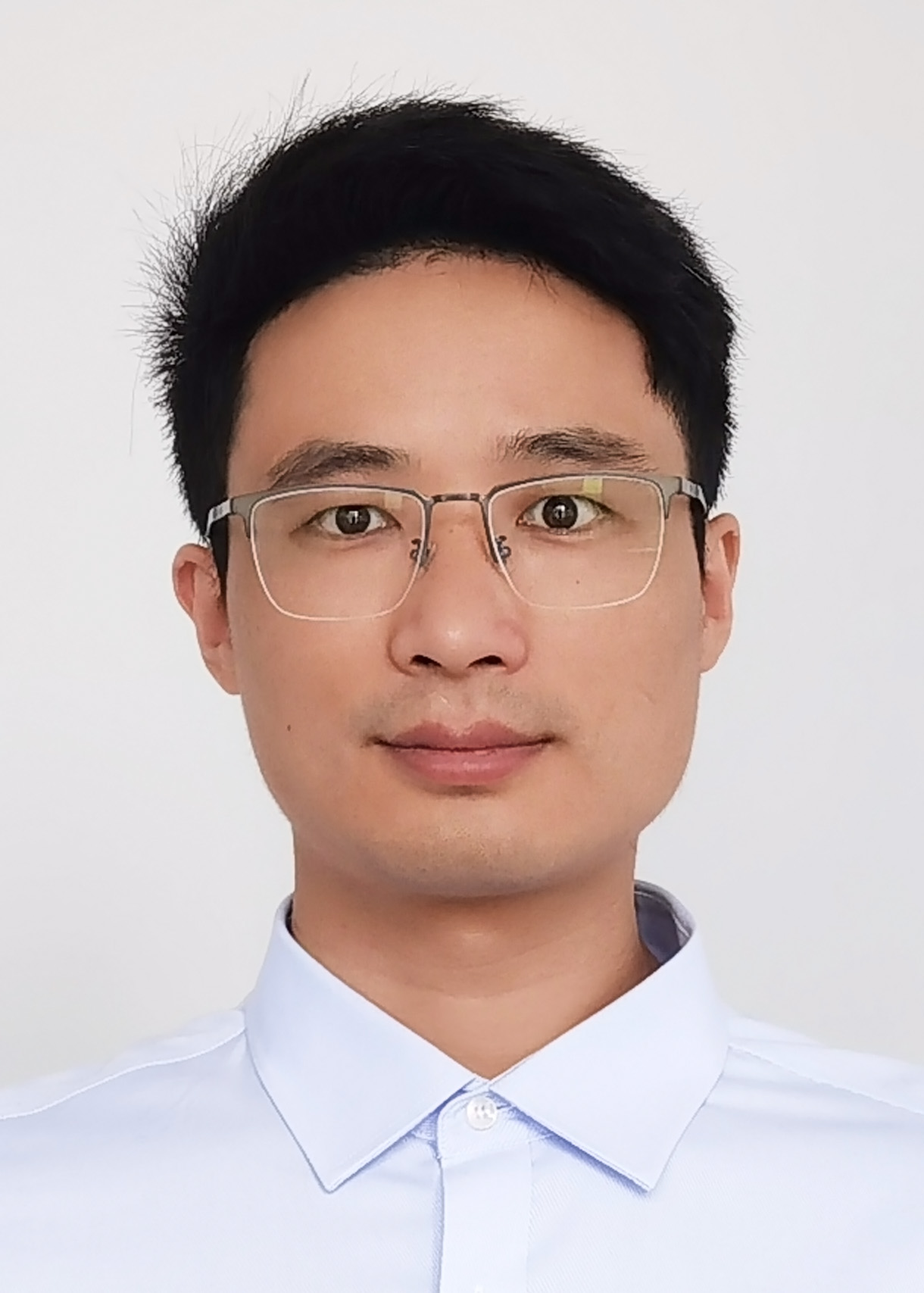}}]{Chongde Zi} received BS and MS degrees from Yunnan Normal University, Kunming, China, in 2014 and 2017, respectively, and the PhD degree from Nanjing University, Nanjing, China, in 2024. He is currently an assistant researcher at Nanjing University, Nanjing, China. His research interests include computational photography and spectral imaging.
\end{IEEEbiography}
\vspace{11pt}

\begin{IEEEbiography}[{\includegraphics[width=1in,height=1.25in,clip,keepaspectratio]{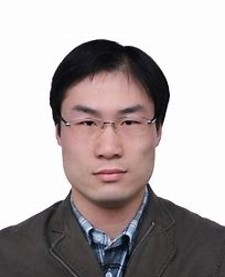}}]{Shuming Wang}
received the BS and Master degree from Suzhou University, China, in 2006, and the PhD degree in physics from Nanjing University, China, in 2009. He is currently a professor at the National Laboratory of Solid State Microstructures, School of Physics, Nanjing University, specializing in nanophotonics, metasurfaces (metamaterials), plasmonics, and quantum optics.
\end{IEEEbiography}
\vspace{11pt}

\begin{IEEEbiography}[{\includegraphics[width=1in,height=1.25in,clip,keepaspectratio]{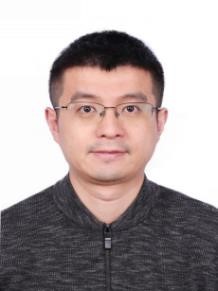}}]{Xun Cao}
(Member, IEEE) received the BS degree from Nanjing University, Nanjing, China, in 2006, and the PhD degree from the Department of Automation, Tsinghua University, Beijing, China, in 2012. He held visiting positions with Philips Research, Aachen, Germany, in 2008, and Microsoft Research Asia, Beijing, from 2009 to 2010. He was a visiting scholar with The University of Texas at Austin, Austin, Texas, from 2010 to 2011. He is currently a professor with the School of Electronic Science and Engineering, Nanjing University. His research interests include computational photography, image-based modeling and rendering, and VR/AR systems.
\end{IEEEbiography}
\vspace{11pt}

\vfill

\end{document}